%% file: neurips_2026.tex
\documentclass{article}

\PassOptionsToPackage{numbers, compress}{natbib}
 \usepackage[preprint]{neurips_2026}

\usepackage[utf8]{inputenc} 
\usepackage[T1]{fontenc}    
\usepackage{hyperref}       
\usepackage{url}            
\usepackage{booktabs}       
\usepackage{amsfonts}       
\usepackage{nicefrac}       
\usepackage{microtype}      
\usepackage{xcolor}         
\usepackage[dvipsnames]{xcolor}
\usepackage[table]{xcolor}
\usepackage{microtype}
\usepackage{graphicx}
\usepackage{subfigure}
\usepackage{subcaption}
\usepackage{enumitem}
\usepackage{booktabs} 
\usepackage{latexsym}
\usepackage{amsmath, amsfonts, amssymb}
\usepackage{tcolorbox}
\usepackage{algorithm}
\usepackage{wrapfig}
\usepackage{array}
\usepackage{algpseudocode}
\usepackage{soul}

\title{Calibrate-Then-Act: Cost-Aware Exploration\\in LLM Agents}

\author{%
  Wenxuan Ding \quad \quad
  Nicholas Tomlin \quad \quad
  Greg Durrett \\
  New York University \\
  \texttt{wd2403@nyu.edu}
}

\begin{document}

\maketitle

\begin{abstract}
  LLM agents are deployed in environments where they must interact to acquire information. In these scenarios, the agent must reason about inherent cost-uncertainty tradeoffs in how to act, such as when to stop exploring and commit to an answer. For instance, on a programming task, an agent might run the code it generates, or it might generate tests for that code snippet; the cost of writing and running a test is nonzero, but typically lower than the cost of running buggy code. 
In this work, we show that we can induce LLM agents to explicitly reason about balancing these cost-uncertainty tradeoffs, then act more optimally in their environments. 
We formalize multiple tasks, including retrieval-augmented QA and a file reading coding task, as sequential decision-making problems under uncertainty. 
Each problem has latent environment state that impacts the agent's performance. We introduce a framework called Calibrate-Then-Act (CTA), where we pass the agent an inferred prior about this environment state to enable it to act more optimally. This information qualitatively changes agent behavior, and adds environment sensitivity to the agent which is not learned via standard RL training.
Our results on a synthetic task, QA, and file reading show that making cost-benefit tradeoffs explicit with CTA helps agents discover more optimal decision-making strategies. 
\end{abstract}

\section{Introduction}
\begin{wrapfigure}{r}{0.53\textwidth}
    \vspace{-35pt}
    \includegraphics[width=\linewidth]{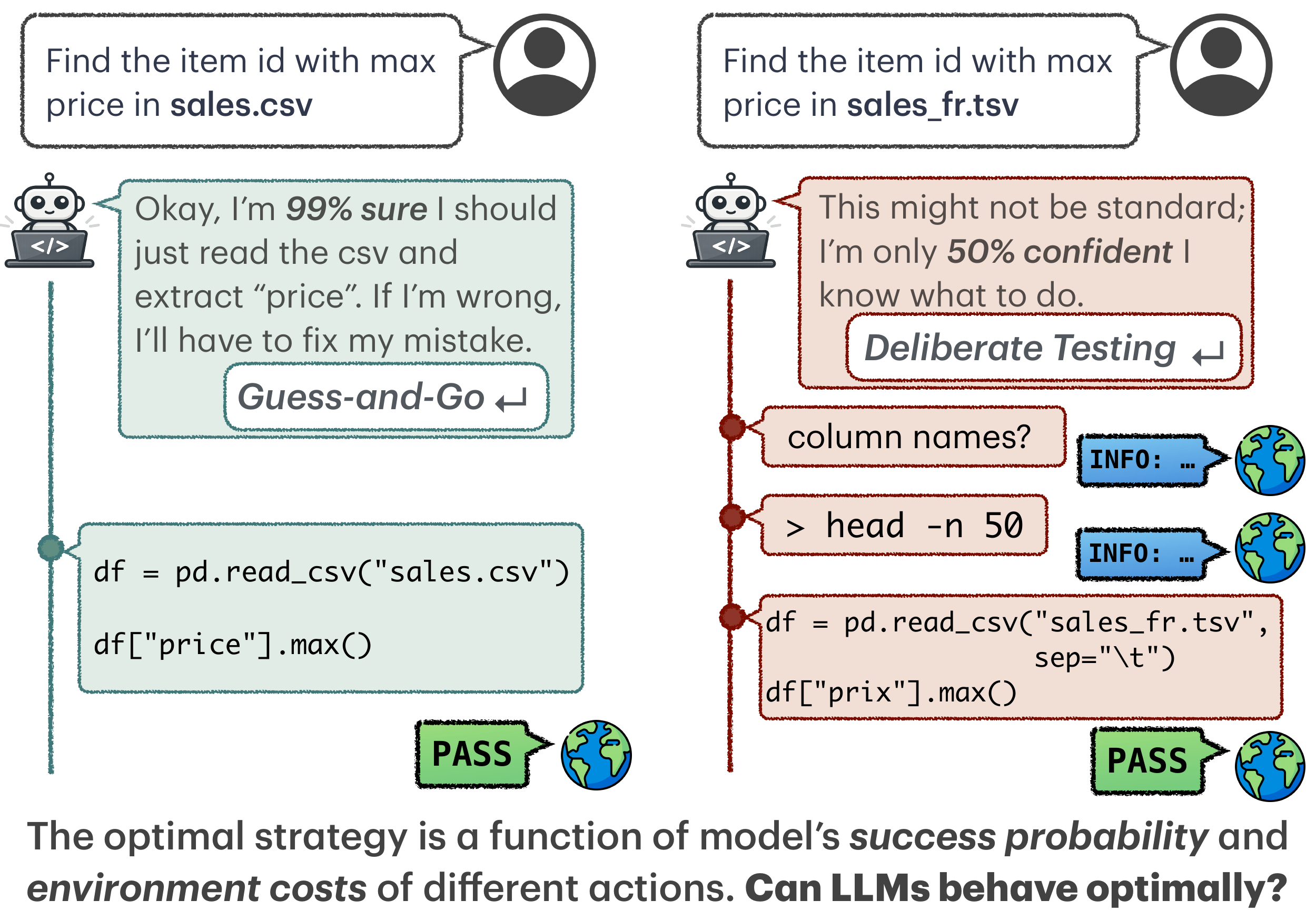}
    \caption{Given the same task, a coding agent may either verify assumptions via intermediate checks carefully (right) or attempt a direct solution as soon as possible (left). The optimal choice depends on uncertainty and specific cost constraints. Calibrate-Then-Act (CTA) materializes this information for better decision-making.}
    \vspace{-15pt}
    \label{fig:intro-code}
\end{wrapfigure}

Large language model (LLM) agents are increasingly tasked with operating in environments where information is incomplete. Behaving rationally requires gaining information by exploring the environment. However, exploration comes with a cost: every additional step increases API costs, interaction latency, and user burden. 

This exploration and its cost come in many forms. In software development and debugging, agents must decide whether to run targeted checks or perform full execution before committing to a solution \citep{zhou2025creditbudgeted}. In machine learning experimentation, practitioners balance inexpensive proxy evaluations against costly full training runs under limited compute budgets \citep{ji2007cost,hennig2024towards,xu-etal-2025-ecotune}. In diagnosis and scientific discovery, additional tests or experiments reduce uncertainty but incur monetary, temporal, or safety costs \citep{karkkainen2019cost,pmlr-v258-li25h,gupta2025llms}. Online decision-making settings such as shopping \citep{yang2023auto,Wang2025RAGENUS}, recommendation \citep{herlihy2024overcoming}, and tool-augmented question answering \citep{yao2023react} exhibit the same structure, as agents weigh further information gathering against acting with partial information. 


Agent policies for this exploration depend in a complex way on their prompt, their inputs, and their training data. However, these policies are frequently \emph{static}. For instance, ChatGPT Deep Research always asks a single round of clarifying questions before searching \citep{Deng2025InteractCompES}, and coding agents like SWE-agent \citep{NEURIPS2024_5a7c9475} start by reading through an existing codebase. This contrasts with settings like Figure~\ref{fig:intro-code}, where we see that a model can act more efficiently if it has confidence that it understands the problem setup and appropriately trades off exploration against action costs.


In this work, we ask: \textbf{how can we develop LLMs that explore in an optimal way under varying cost and uncertainty profiles?} Leveraging the strong reasoning capabilities of modern LLMs, we propose a framework called Calibrate-Then-Act (CTA), which decouples the calibration of uncertainty from the reasoning of action selection. The key insight is that by presenting priors explicitly to the model, we induce the model to reason about an underlying sequential decision-making problem abstractly and discover the optimal action.

We study two settings inspired by realistic problems: (1) knowledge-based QA with optional retrieval, where uncertainty is inferred from the model's own confidence; and (2) coding tasks where priors regarding environmental structure (e.g., file schemas) are derived from cues learned through past experience. We show that even in these simple settings, LLM agents prompted or trained with RL fail to behave optimally. However, this is not due to an inability to perform the correct reasoning: we show with a synthetic ``Pandora's Box'' problem \citep{pandora}, 
and show that even a small thinking model (Qwen3-8B) can follow the optimal policy. The Calibrate-Then-Act method builds on this result, feeding the agent with relevant information about the environment, determined by either estimating model confidence or a trained predictor of environment state. A zero-shot prompted implementation of this method, \textsc{CTA-Prompted}, already enables dynamic decision-making. We show that training with RL further improves performance, and notably outperforms baseline RL training of the agent.

Our contributions are: (1) Framing environment exploration as a sequential decision-making problem, unified across three settings; (2) The Calibrate-Then-Act method, which induces LLMs to reason about the optimality of their actions and achieve better cost-performance tradeoffs than baselines.





\section{Background: Agentic Exploration with Costs}
\label{sec:unified-form}




We formalize cost-aware environment exploration task as sequential decision-making problem. Our agent, which for the purposes of this work is an LLM, is given a query $\mathbf{x}$ and operates in some environment, which can be defined as a partially-observable Markov Decision Process $\mathcal{W} = (\mathcal{S}, \mathcal{A}, \mathcal{O}, O, T, R, D_\theta)$, a tuple of states $\mathcal{S}$, actions $\mathcal{A}$, observations $\mathcal{O}$, observation function $O$, transition function $T$, reward function $R$, and parameterized discount function $D_\theta$, which integrates the cost.

In the settings we consider, $\mathcal{A}$ and $\mathcal{O}$ are both string-valued spaces; LLMs produce string actions (code, API calls, etc.) and receive string-valued responses from the environment. The observation function $O$ produces string realizations of the underlying environment; e.g., in Figure~\ref{fig:intro-code}, the results of executing commands return string output in the terminal reflecting the underlying state of the environment.

The environment contains problem-critical unobserved features that will determine the agent's performance, e.g., details about the formatting of the unobserved file in Figure~\ref{fig:intro-code}. These can be thought of as a subset of the information in $\mathcal{S}$. We represent these as a random variable $Z$ taking values $\mathbf{z} \in \mathcal Z$.



The agent interacts with the environment over multiple timesteps before terminating.
At each timestep $t$, the agent selects an action $a_t \in \mathcal A$ and receives an observation $o_t \in \mathcal O$; for simplicity, we assume that $o_t$ encodes $a_t$. Based on this information, we can form an idealized posterior distribution $b_t(Z) = p(Z \mid \mathbf{x}, o_{0:t})$ which reflects remaining uncertainty over the latent variables. 
The action space generally consists of multiple \emph{exploration} actions and a \emph{commit} action, which terminates the episode by producing a final result.
In Figure~\ref{fig:intro-code}, exploration actions include querying aspects of the input data file. Other actions (not shown) include running the code or writing and running unit tests. 

Each action incurs different costs depending on the setting, while the commit action corresponds to returning a final solution. These costs are reflected by a discount factor  $D_\theta(a_{1:T}) \in [0,1]$, which discounts the value of successful task completion based on the exploration actions taken prior to commitment.

Finally, the agent receives reward upon committing:
\[
R = \mathbb{I}[\text{task completed at $a_t$}] \cdot D_\theta(a_{1:T}),
\]

Overall, the agent’s objective is to maximize the expected discounted reward $R$ by carefully selecting actions that adaptively balance exploration and commitment in response to uncertainty and cost constraints of the environment. 

\section{Tasks}

We now connect our general task setting to three concrete problems we address in this work. The first, Pandora's Box~\citep{pandora}, serves as a proof-of-concept instance of cost-aware exploration, where we demonstrate that even small LLMs have the capability to reason about uncertainty-cost tradeoffs in environments. 
We then describe two realistic problems where LLMs struggle to reason about costs out of the box, motivating the method in this paper.
Table~\ref{tab:symbol-mapping} summarizes how these problems map onto our general framework.

\begin{table*}[t]

\input{tables/symbol-mapping}
\end{table*}

\subsection{Pandora’s Box: A Minimal Setting for Cost-Aware Decisions}
\label{sec:poc}




\paragraph{Formalization}
We consider a variant of the classic Pandora's Box Problem with discounted reward over time~\citep{pandora}. 
The task involves a finite set of $K$ boxes $\{z_1, z_2,\ldots,z_K\}$, among which exactly one box $z^*$ contains a prize of value $1$. This box is unknown to the agent and is drawn from a prior distribution 
\[p(z_k = z^*) = p_k, \quad \sum_{k=1}^{K} p_k = 1.\]
At each timestep $t$, the agent can either \emph{verify} a box $z_{k_t}$ of its choice to check if it contains the prize and discount their final reward by $\gamma \in [0,1]$, or \emph{commit} to a box given its current information and receive a reward $R=\gamma^{t} \cdot \mathbb{I}(z_{k_t} = z^{*}).$

This task can be viewed as a toy version of real-world use cases for LLM agents. For instance, if trying to identify a bug in a piece of code, we can view committing as directly taking action to fix a bug and verifying as writing a unit test to check the correctness of the code.


Knowledge of the prior probabilities is necessary to behave optimally.
The optimal policy is to commit if its posterior probability of a box is greater than $\gamma$, which means we have better expected value for guessing than for verifying. Otherwise, the optimal decision is to verify the highest-probability remaining box. The full algorithm is shown in Algorithm~\ref{alg:pandora-k3} and proof is provided in Appendix~\ref{app:oracle-proof}. 



\subsection{Task~\textsc{QA}: Knowledge QA with Optional Retrieval}
\label{sec:qa-form}
We study a knowledge question answering setting in which an LLM can optionally acquire external information at a cost \citep{eisenstein2025dont}.
Given a factual query, the model must decide whether to rely on its parametric knowledge or defer commitment and retrieve additional evidence, trading off potential accuracy gains against latency and API costs.


\paragraph{Formalization}
Given a question $\mathbf{x}$, a discount factor $\gamma \in [0,1]$, and access to a retriever, the agent chooses between actions $\mathcal{A} =\{\mathrm{retrieve}, \mathrm{answer}\}$.  
A retrieve action queries a retrieval system with the input $\mathbf{x}$, while an answer action invokes the LLM to answer the question given the context so far, producing an answer $a$. The model receives reward $R = \gamma^{t} \cdot \mathbb{I}(a = a^{*})$, where $a^{*}$ is the ground-truth answer to $\mathbf{x}$. This task only features up to one round of retrieval, so $t$ is either 0 or 1.

\paragraph{Latent structure and prior} There are two relevant latent variables for this problem. First, we define $p_\text{da} = p(a = a^{*} \mid \mathbf{x})$ that the model will correctly answer the question if asked directly, without retrieval. We model $p_\text{da}$ with a distribution $p(p_\text{da} \mid \mathbf{x}) = \delta(z=k(\mathbf{x}))$ for an estimate $k_\text{da}(\mathbf{x})$ that the LLM returns the correct answer, where $\delta$ is the Dirac delta function. This distribution represents that we have a point estimate of the model's probability to answer correctly.

Second, we define the probability of answering post-retrieval to be $p_\text{ret} = p(a'=a^* \mid \mathbf{x}, \mathbf{c})$, where $\mathbf{c}$ is the retrieved context. We similarly represent this as a delta function $p(p_\text{ret} \mid \mathbf{x}) = \delta(z=k_\text{ret})$

Under this abstraction, the oracle policy retrieves whenever the expected discounted accuracy after retrieval exceeds that of direct answering: $p_{\text{ret}} \cdot \gamma \;\ge\; p_{\text{da}}$.

\subsection{Task~\textsc{FileReading}: Coding with Selective Testing}
\label{sec:code-form}

Following Figure~\ref{fig:intro-code}, 
we design a coding setting to encapsulate the idea of a model choosing to either run code directly or test it. 
Concretely, the model needs to write code to load a data file with correct formatting options and return the answer to a task query (e.g., identifying the \texttt{user\_id} associated with the maximum \texttt{score}). The true file schema is not specified in the input; instead, the model may infer likely formats from filename cues and past experience. Unlike the QA setting, these priors are not known to the model a priori, but must be learned from training on this task. Despite the simplicity of this setting, we show that LLM agents fail to learn the optimal behavior via RL. 
\paragraph{Formalization}
Given a query $\mathbf{x}$ that contains a task specification and a CSV filename $n$, the agent needs to write code to load the file correctly and compute an answer $a$. Specifically, each file is associated with latent formatting attributes $\mathbf z = (z_d, z_q, z_s) \in \mathcal Z$, where $z_d \in \{\texttt{,}, \texttt{;}, \texttt{\textbackslash t}\}$ denotes the delimiter,
$z_q \in \{\texttt{'}, \texttt{"}\}$ the quote character, and
$z_s \in \{0,1\}$ the number of skipped header rows. Without making a correct inference about $\mathbf{z}$, the task is not solvable.

At each timestep $t$, the agent selects one action $a_t \in \mathcal A$ from three types:
\texttt{UNIT\_TEST}($f$), \texttt{CODE}($d, q, s$), or \texttt{ANSWER}.
A \texttt{UNIT\_TEST} probes a chosen formatting attribute $f$ and reveals its true value.
A \texttt{CODE} action executes code written under the agent’s current belief ($d,q,s$) and returns feedback via \texttt{stdout} and \texttt{stderr}, which may contain the answer or signals useful for debugging and refinement.
An \texttt{ANSWER} action commits to a final answer $a'$ and terminates the episode.
The agent may interleave \texttt{UNIT\_TEST} and \texttt{CODE} actions in any order, and may perform multiple \texttt{CODE} actions to refine its solution based on previous execution feedback.

Each \texttt{UNIT\_TEST} and \texttt{CODE} action incurs multiplicative discounts $d_u$ and $d_c$, respectively.
Upon committing at time $T$ by \texttt{ANSWER}, the agent receives reward $R = d_u^U \cdot d_c^C \cdot \mathbb{I}(a' = a^*)$, where $a^*$ denotes the ground-truth answer.
\vspace{-8pt}

\paragraph{Prior} The prior distribution $p(\mathbf{z} \mid n)$ over formatting attributes may be inferred from conventions or past experience, or provided explicitly by a format predictor.

Prompt templates for these tasks are provided in Appendix~\ref{sec:prompt-all}.

\section{Method: Calibrating Agent Environment Exploration}
\label{sec:calibrating}

\begin{figure*}[t]
    \centering
    \includegraphics[width=0.95\linewidth]{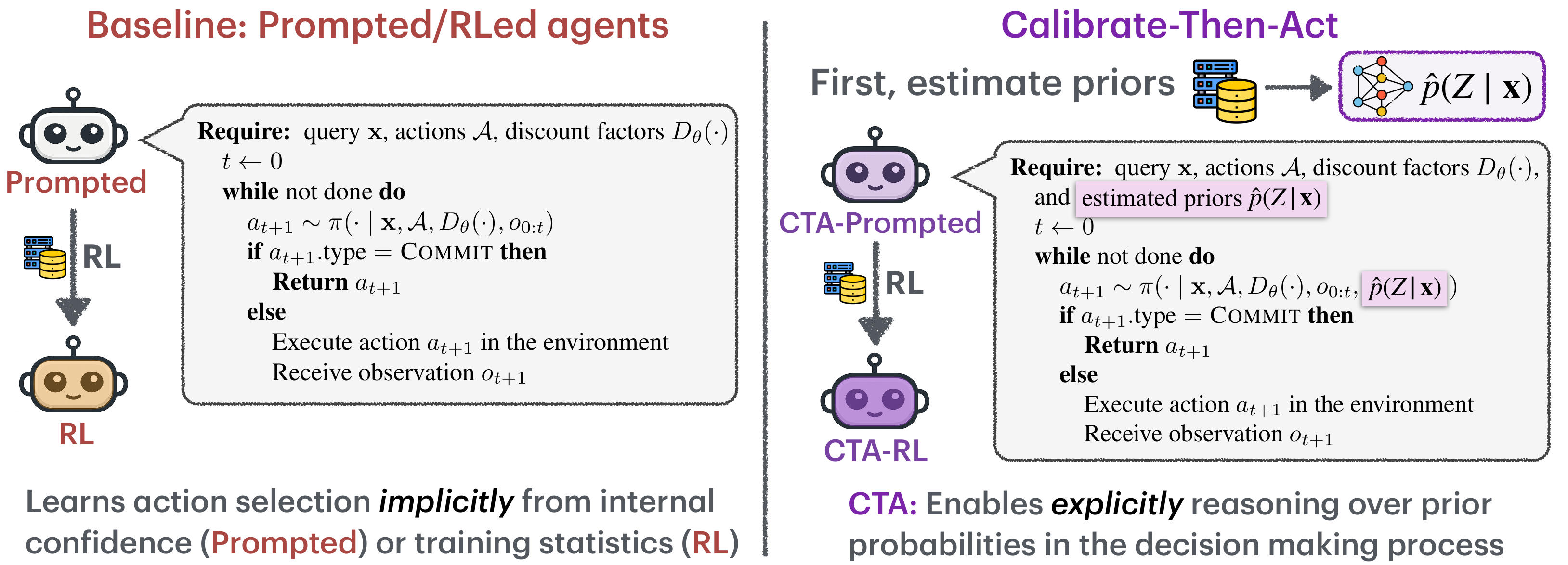}
    \caption{Standard agentic decision loop (left) and proposed method \textsc{CTA} (right). In CTA, we learn a prior estimator from training data and condition the agent on estimated $\hat p$ at inference and/or training time, inducing more optimal decision making through explicit reasoning over prior probabilities.}
    \label{fig:calibrator}
    \vspace{-12pt}
\end{figure*}

To behave Pareto-optimally in a sequential decision-making problem, an agent must jointly compare the cost of additional exploration against the expected value of additional information to decide whether to continue exploring or to commit based on its current partial information. 
The value of additional information depends on reasoning over current beliefs about the underlying world state via the prior $p(\mathbf{z} \mid \mathbf{x})$ and posterior $b_t$.

We define our LLM agent as $\pi(a_t \mid \mathbf{x}, \mathcal{A}, D_{\theta}(\cdot), o_{0:t})$, placing a distribution over the next action in a given state. Figure \ref{fig:calibrator} (left side) shows the basic form of this agent. $\pi$ can be implemented either via a prompted LLM or through a model trained with reinforcement learning. 
However, in practice, it is difficult for $\pi$ to learn to do the right reasoning in the environments we evaluate on. 

Our key methodological contribution is to \emph{explicitly} provide estimates of the prior, denoted as $\hat{p}(Z \mid \mathbf{x})$. Figure \ref{fig:calibrator} (right side) illustrates the role of a prior estimate in the agentic decision loop. 
We instantiate two versions of this approach.

\begin{itemize}[leftmargin=10pt]
\item \textbf{CTA-Prompted:} In this setting, we use the same agent $\pi$ as the baseline, but with a modified prompt to include the prior information $\hat{p}(Z \mid \mathbf{x})$ together with $\mathbf{x}$. Examples of the prompt are given in Appendix~\ref{sec:prompt-all}.
\item \textbf{CTA-RL:} In this setting, we use the prompt of CTA-Prompted and perform additional training via RL. We fine-tune the model end-to-end using GRPO \citep{shao2024deepseekmath} with the discounted reward objective from the task at hand. We evaluate it on new examples with the same prompt used at training time. 
\end{itemize}

Both settings rely on the ability of a thinking LLM to determine how to use this information. In Section~\ref{sec:pandora-results}, we show that the Pandora's Box problem is solvable by thinking models given the appropriate priors. In Section~\ref{sec:cta-results} and Figure~\ref{fig:qa-decision-boundary}, we then show that thinking models also perform substantially better on our tasks of interest.



Having accurate prior estimates is important for CTA to do well. We now describe how to estimate the priors described in Table~\ref{tab:symbol-mapping}. For Pandora's Box, the priors are available explicitly, but for our other two problems, estimation is needed. Our two problems require two different ways of estimating these priors. However, a commonality is that both estimation techniques draw inferences about the problem that are not easily learned from end-to-end RL, thereby motivating the need for CTA.

\subsection{Estimating Priors from Model Confidence (\textsc{QA})}

In \textsc{QA}, the true $k_\text{da}(\mathbf{x})$, or the probability that the model can answer $\mathbf{x}$ correctly without retrieval, is not directly observable. This problem is one of confidence estimation. There are several ways to obtain estimates of model confidence, including inspecting logits, probe-based methods, and verbalized confidence.
In this work, we use verbalized confidence \citep{mohri2024language}, which is a general method that can apply to both blackbox and glassbox model usage. Given a question $\mathbf{x}$, we prompt the model to produce a discrete verbalized confidence value $p_v(\mathbf{x})$, and apply an isotonic regression model \citep{zadrozny2002transforming} $\mathrm{ISO}$ trained on the validation set to obtain a calibrated estimate
\[
\hat{k}_{\text{da}}(\mathbf{x})
 = \mathrm{ISO}\big(p_v(\mathbf{x})\big).
\]
The expected calibration error (ECE) reduces from $0.618$ to $0.029$ on the PopQA dataset \citep{mallen-etal-2023-trust} (described in Section~\ref{sec:experiment-setup}) with this regression. This drop is reflective of initial poor calibration \citep{guo2017calibration,xiong2024can, shen2024smartcal,wang2025self} and demonstrates that rescaling can help \citep{desai-durrett-2020-calibration}. This kind of recalibration is very hard for the agent to learn with approaches like end-to-end RL: if RLed for a long time on a QA dataset, a model is more likely to memorize specific facts than it is to systematically recalibrate its beliefs.


\subsection{Estimating Priors from Training Data (\textsc{FileReading})}
Coding agents bring to bear many kinds of priors when completing tasks. Our \textsc{FileReading} task studies inferences about file formats as a controlled representative problem. In our setting, we assume that the agent is trained on our task, and therefore both baseline RL and the CTA methods are exposed to data showing the mapping of filenames to format information. 

For prior estimation, we train a filename-to-format predictor, denoted as $\mathcal M_{\text{BERT}}$, to estimate the distribution $p(\mathbf z \mid n)$ from the filename. The predictor is based on a lightweight BERT-tiny encoder (4.4M parameters) \citep{bhargava2021generalization, DBLP:journals/corr/abs-1908-08962}.
Given a filename \(n\), the model encodes the tokenized string and uses the \texttt{[CLS]} representation to produce three independent categorical distributions via linear heads: delimiter, quote character, and skiprows.
The model is trained with a summed cross-entropy objective across the three heads for one epoch on the training split. On the validation split, $\mathcal M_{\text{BERT}}$ achieves an average classification accuracy of $67\%$ across the three formatting attributes. This accuracy is expected: the ground-truth format values are sampled from a soft filename-dependent prior whose most likely option has probability about $67\%$ on average. Thus, the task is intentionally stochastic rather than deterministically predictable from the filename. Appendix~\ref{app:prior_sensitivity} shows that CTA performs worse as this predictor gets worse, but in a smooth fashion, with noisy predictors still giving benefit.

After training, $\mathcal M_{\text{BERT}}$ outputs marginal probabilities $\{p(z_d \mid n), p(z_q \mid n), p(z_s \mid n)\}$, which are provided to the agent associated with each task during RL training or test time, thereby decoupling uncertainty estimation from action selection.


\section{Experimental Setup}
\label{sec:experiment-setup}
\paragraph{Datasets}

For \textsc{QA}, we evaluate on PopQA~\citep{mallen-etal-2023-trust}, a QA benchmark that covers long-tail factual knowledge and benefits from retrieval. We sample 1{,}000 questions for evaluation and build the retriever based on \textbf{Contriever} \citep{izacard2022unsupervised,bajaj2016ms}.
For each question, we sample a discount factor $\gamma \sim \mathcal{U}[0.1, 0.65]$ to study model behavior across varying retrieval costs.

Our coding problem involves a CSV-based question-answering dataset called \textsc{FileReading}, where filename cues provide informative signals about file formats and correct parsing requires executing code with appropriate format values. At test time, the true file format is hidden and only the filename is provided as part of the task query $\mathbf{x}$. 
\textsc{FileReading} contains 2{,}000 tasks, split into 1{,}400 training, 300 validation, and 300 test examples.
For each task, we randomly sample a unit-test discount $d_u$ from $[0.5, 1]$ and duplicate the instance across four code discount settings $d_c = d_u^{\rho}$, with $\rho \in \{0.5, 1.0, 2.0, 4.0\}$, varying the relative cost of code execution while holding the task fixed. Details of the dataset construction are provided in Appendix \ref{app:code-data-generation}.


\paragraph{Metrics}
We evaluate the model performance across three sets of metrics. (1) Exploration statistics: for QA, we report retrieval rate $\text{Retrieve}\%$ (the fraction of questions for which retrieval is invoked), and for \textsc{FileReading}, we report the number of unit tests $U$ and code attempts $C$. (2) Task accuracy, which measures whether the final model output matches the ground-truth answer for a task query. (3) Reward, which discounts correctness with exploration costs. 

\paragraph{Models and Baselines}

We use Qwen3-8B \citep{qwen3technicalreport} as the base model for both the \textsc{QA} and \textsc{FileReading} tasks. We compare the following baselines against our methods \textsc{CTA-Prompted} and \textsc{CTA-RL}.
\begin{itemize}[leftmargin=20pt,itemsep=0pt, topsep=0pt]
    \item \textsc{Prompted}: We prompt the base model directly with task description and query $\mathbf{x}$.
    \item \textsc{Prompted-NonThink} : Similar to \textsc{Prompted}, but with thinking mode disabled by prepending \texttt{<think></think>} tags without thinking content. Unless specified with \textsc{-NonThink}, we enable thinking mode by default for all other settings.
    \item \textsc{RL}: We fine-tune the model end-to-end using GRPO \citep{shao2024deepseekmath} with discounted reward objective, and evaluate it by prompting with the task description and query $\mathbf{x}$.
\end{itemize}

To expose both RL settings to diverse cost trade-offs in \textsc{FileReading}, we duplicate each training instance across multiple relative cost values $\rho \in \{0.5, 1.0, 2.0, 4.0\}$, yielding a $4\times$ larger RL training set than that used for training $\mathcal{M}_{\text{BERT}}$. More training details are in Appendix~\ref{app:grpo-training-details}.

\vspace{-5pt}
\section{Results}
\vspace{-5pt}


\begin{figure*}[t!]
    \centering
    \vspace{-20pt}
    \includegraphics[width=0.95\linewidth]{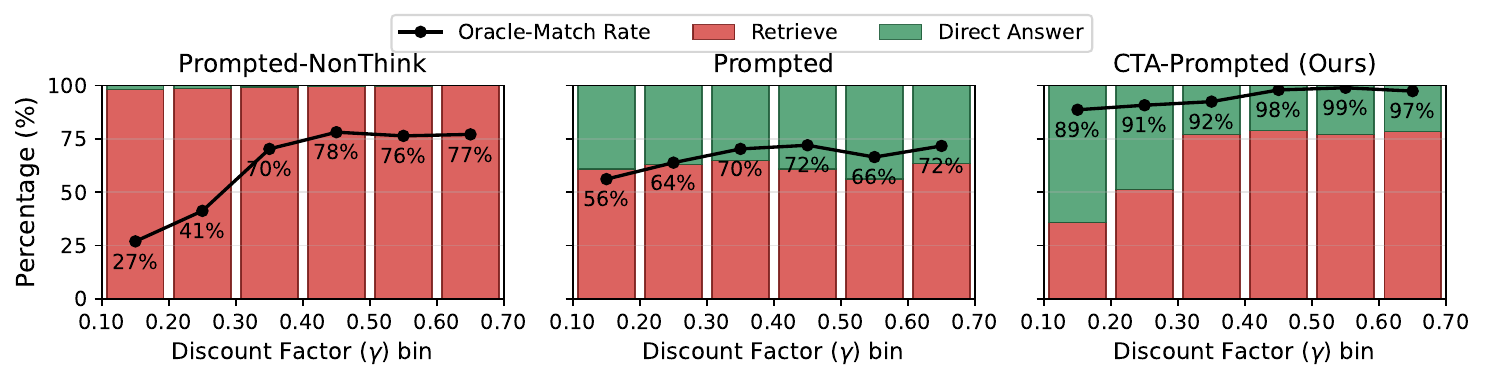}
    
    \caption{Model retrieval decision across retrieval discount factors $\gamma$ on the \textsc{QA} task. 
    Red bars show the percentage of examples where the model retrieves before answering; green bars show direct answers.
The black line reports the oracle policy match rate. 
\textsc{CTA-Prompted} makes adaptive cost-sensitive decisions and achieves highest oracle match rates across settings.}
 \vspace{-12pt}
    \label{fig:qa-decision-boundary}
\end{figure*}

\subsection{LLMs can solve Pandora's Box problems}
\label{sec:pandora-results}
\begin{wraptable}{r}{0.53\linewidth}
\centering
\vspace{-12pt}

\caption{Performance comparison on the Pandora's Box task. A thinking model given access to priors can solve the problem near-optimally.} 
\resizebox{1\linewidth}{!}{
\begin{tabular}{l c >{\columncolor{gray!15}}c}
\toprule[1.5pt]
Method & Optimal Match Rate (\%) & Avg. Reward \\
\midrule[0.75pt]
Oracle policy & 100.0 & 0.649 \\
\midrule[0.75pt]
\textsc{Prompted-NonThink} & 11.0 & 0.441 \\
\textsc{Prompted} & 23.0 & 0.476\\
\textsc{CTA-Prompted-NonThink} & 20.0 & 0.436 \\
\textsc{CTA-Prompted}  & \textbf{94.0}& \textbf{0.625} \\
\bottomrule[1.5pt]
\end{tabular}}
\vspace{-20pt}
\label{tab:pandora-results}
\end{wraptable}
We first evaluate whether LLM agents can solve the proof-of-concept Pandora's Box task introduced in Section~\ref{sec:poc}.
We instantiate the agent using the framework in Figure~\ref{fig:calibrator}, with prompts in Figure~\ref{fig:prompt-pandora}, and use Qwen3-8B as the base model. We evaluate on $100$ examples with $K =3$, discount factors sampled from $[0,0.1,\ldots, 1.0]$, and priors sampled from a symmetric Dirichlet distribution with concentration parameter $\alpha=0.5$.

 \begin{wrapfigure}{r}{0.53\textwidth}
 \vspace{-15pt}
    \includegraphics[width=\linewidth]{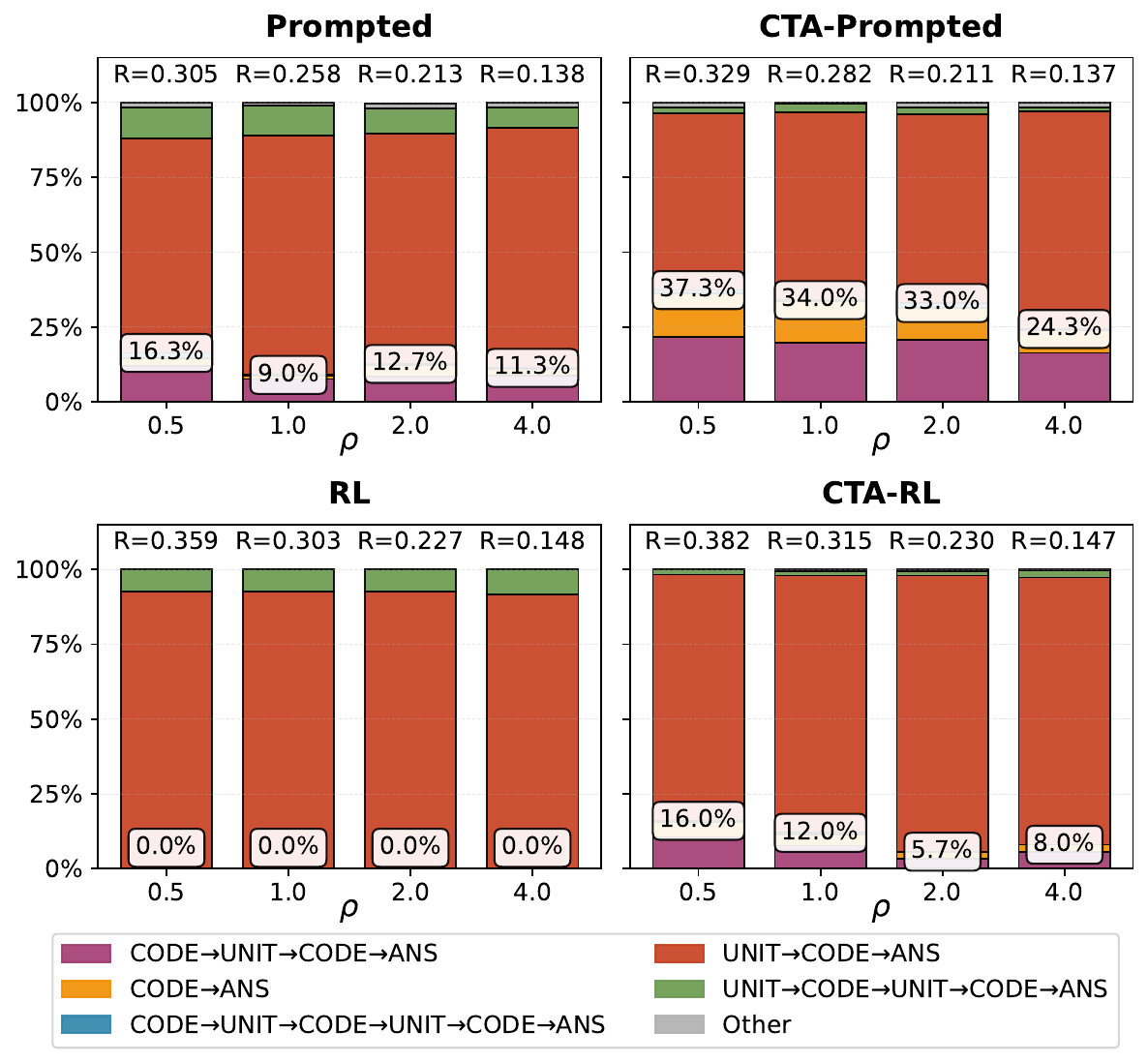}
    \caption{Action distribution for agents on \textsc{FileReading} across relative cost parameters $\rho$. Each stacked bar shows the proportion of decision traces corresponding to different action patterns, with the reward $R$ labeled above. Annotated percentages indicate the fraction of tasks where the agent attempts code execution before any unit tests.}
    \vspace{-24pt}
    \label{fig:breakdown-by-rho}
\end{wrapfigure}

Table~\ref{tab:pandora-results} reports the performance with two metrics: average reward and optimal policy match rate, measuring whether the model’s interaction trace aligns with the oracle strategy described in Appendix  Algorithm~\ref{alg:pandora-k3}.
Without explicit priors or when the thinking mode is disabled, the agent exhibits low optimal match rates, indicating failure to recover the optimal decision rule. In contrast, \textsc{CTA-Prompted} achieves $94.0\%$ optimal match rate and significantly higher reward, 
indicating that the model is capable of reasoning about optimal exploration-commitment tradeoffs and adapting to different discount factors and prior distributions when the environment constraints are made explicit.
Appendix~\ref{app:pandora-trace-analysis} compares interaction traces across settings, showing how explicit reasoning in thinking traces is used. \textbf{Crucially, the presence of explicit prior information triggers the model to reason about cost-accuracy tradeoffs, from which it can make optimal decisions.} 

\subsection{Calibrate-Then-Act enables models to make calibrated decisions}
\label{sec:cta-results}

\paragraph{Baseline LLM agents struggle to make calibrated decisions, even after RL.}

Across both \textsc{QA} and \textsc{FileReading}, we find that without explicit priors, LLMs do not naturally form calibrated decision rules from task descriptions alone.


For \textbf{\textsc{QA}}, Figure~\ref{fig:qa-decision-boundary} visualizes the retrieval behavior across retrieval discount factors $\gamma$. 
Each stacked bar shows the model's decisions for problems within a discount factor bin: \textcolor{red}{red} denotes the percentage of examples where the model retrieves before answering, while \textcolor{ForestGreen}{green} denotes direct answers. The black line reports oracle match rate, i.e., the fraction of decisions that match the optimal retrieve-or-not rule defined by the threshold $k_{da} = \gamma \cdot k_{ret}$. Specifically, \textsc{Prompted-NonThink} (Figure~\ref{fig:qa-decision-boundary}, left) retrieves almost always in all cost regimes, reflecting a conservative and cost-insensitive strategy.

Enabling thinking reduces unnecessary retrieval, but \textsc{Prompted} (Figure~\ref{fig:qa-decision-boundary}, mid) adapts only weakly to retrieval costs and achieves modest oracle-match rates. 
Although \textsc{Prompted} sometimes reasons qualitatively about its parametric knowledge, it rarely quantifies these as certainty values for an expected-value comparison. Similarly, it sometimes mentions costs, but does not consistently tie the final decision to the costs.
As a result, its decisions often fall back to a recall-based heuristic: answer if a plausible candidate comes to mind, retrieve otherwise, rather than systematically adapting to $\gamma$ and confidence.
This suggests that, without explicit prior estimates, the models struggle to assess their own parametric knowledge and make consistent retrieval decisions.

On \textbf{\textsc{FileReading}}, Figure~\ref{fig:breakdown-by-rho} shows a similar analysis. The x-axis $\rho = \log{d_c} /\log{d_u}$ reflects relative cost: as $\rho$ increases, code attempts become relatively more costly, and a rational agent will tend to test more before coding.
Each stacked bar represents the collection of tasks with a specific $\rho$, and each color represents the proportion of action trace pattern in the collection. The reward is labeled above each bar, and the percentage of ``guess-and-go'' \tcbox[
  on line,
  colback=gray!5,
  colframe=black,
  boxrule=0.6pt,
  arc=3pt,
  left=3pt,
  right=3pt,
  top=0.5pt,
  bottom=0.5pt
]{x\%} (attempting code without any preceding unit tests) are also labeled.

Both \textsc{Prompted} and \textsc{RL} collapse to a static test-first policy rather than adapting across cost regimes, with \tcbox[
  on line,
  colback=gray!5,
  colframe=black,
  boxrule=0.6pt,
  arc=3pt,
  left=3pt,
  right=3pt,
  top=0.1pt,
  bottom=0.1pt
]{0\%} of ``code-first'' traces (attempts to execute code before any unit tests). RL without prior cannot internalize the structure of the training data and instead defaults to a suboptimal exploration policy. Together, these results show that LLMs struggle to make calibrated, cost-sensitive decisions when prior estimates are implicit.

\begin{table*}[t!]
\centering
\begin{minipage}{0.47\textwidth}
\centering
\small
\caption{Performance on \textsc{QA}. Our principal metric is reward. One-turn baselines use fixed strategies, while multi-turn agents adaptively decide when to retrieve. \textsc{CTA-Prompted} achieves the highest reward.}
\centering
\resizebox{\linewidth}{!}
{%
\begin{tabular}{l c c >{\columncolor{gray!15}}c}
\toprule[1.5pt]
Method & Retrieve \% & Acc. & Reward \\
\midrule[0.75pt]
\multicolumn{4}{l}{\emph{Single-turn baselines}} \\
\textsc{Never Retrieve} & 0.0 & 0.226 & 0.226 \\
\textsc{Always Retrieve} & 100.0 & 0.578 & 0.213 \\
\midrule[0.75pt]
\multicolumn{4}{l}{\emph{Multi-turn agents}} \\
\textsc{Prompted-NonThink} & 97.7 & 0.619 & 0.244 \\
\textsc{Prompted} & 61.4 & 0.501 & 0.283 \\
\textsc{CTA-Prompted} (Ours) & 65.3 & 0.512 & \textbf{0.293} \\
\bottomrule[1.5pt]
\end{tabular}

}
\vspace{-12pt}
\label{tab:qa-retrieval-results}
\end{minipage}
\hfill
\begin{minipage}{0.50\textwidth}
 \vspace{-15pt}

\caption{Performance on \textsc{FileReading}, averaged across relative unit-test and code-execution cost ratios $\rho \in \{0.5, 1.0, 2.0, 4.0\}$. We report the average number of turns, unit-test calls (U), code executions (C), accuracy, and discounted reward.}\small
\resizebox{\linewidth}{!}
{
\begin{tabular}{l c c c c >{\columncolor{gray!15}}c}
\toprule[1.5pt]
\textbf{Method} & \textbf{\# Turns} & \textbf{U} & \textbf{C} & \textbf{Acc.} & \textbf{Reward} \\
\midrule[0.75pt]

\multicolumn{6}{l}{\textit{Without Training}} \\
\textsc{Prompted}   & 3.62 & 2.67 & 1.42 & 0.958 & 0.229 \\
\textsc{CTA-Prompted} (Ours)   & 3.47 & 2.51 & 1.41 & 0.945 & 0.240 \\

\midrule[0.75pt]
\multicolumn{6}{l}{\textit{With RL Training}} \\
\textsc{RL}   & 3.51 & 2.13 & 1.39 & 0.997 & 0.259 \\
\textsc{CTA-RL} (Ours)  & 3.46 & 1.98 & 1.46 & 0.991 & \textbf{0.268} \\

\bottomrule[1.5pt]
\end{tabular}

}

\vspace{-10pt}
\label{tab:qwen8b-code-performance}
\end{minipage}
\end{table*}

\paragraph{CTA pushes the model towards adaptive, optimal decision making in both task settings.} Our main results are given in Tables~\ref{tab:qa-retrieval-results} and \ref{tab:qwen8b-code-performance}. 
Table~\ref{tab:qa-retrieval-results} compares \textsc{QA} performance across fixed retrieval strategies and various multi-turn settings.
\textsc{
CTA-Prompted} achieves the highest reward, outperforming both \textsc{Prompted-NonThink}, which retrieves too often ($97.7\%$), and \textsc{Prompted}, which reduces retrieval but does not align as well with the oracle policy. Figure~\ref{fig:qa-decision-boundary} shows that CTA-Prompted is much more adaptive in its behavior than baselines: it retrieves much more frequently at high discount factors and less at low discount factors. This is also reflected in it obtaining the highest oracle match rate across cost constraints.

Table~\ref{tab:qwen8b-code-performance} shows the model performance on \textsc{FileReading} aggregated across different cost settings. 
Note that both \textsc{RL} and  \textsc{CTA-RL} have access to the same set of training data which encodes correlations between filenames and file formats implicitly; \textsc{CTA-Prompted} uses more information than \textsc{Prompted} by virtue of the trained model.
The agent achieves a discounted reward of $0.259$ by training end-to-end (\textsc{RL}) with the discounted reward as objective, while conditioning the training on explicit estimated priors (\textsc{CTA-RL}) further improves by  $3.5\%$ with an overall reward of $0.268$. This shows that incorporating estimated priors helps the model generalize better on unseen test data than end-to-end RL.

Figure~\ref{fig:breakdown-by-rho} 
shows that while end-to-end RL improves reward over its non-training counterpart \textsc{Prompted}, 
its decision-making behavior collapses to a static policy which always triggers unit tests before the first code attempt ($0\%$ ``guess-and-go''). 
In contrast, \textsc{CTA-Prompted}, which conditions the agentic decision-making on estimated priors, already exhibits adaptive behavior in response to costs by acting more conservatively with higher $\rho$. After training, this adaptivity remains pronounced in \textsc{CTA-RL}.

\begin{wrapfigure}{r}{0.53\textwidth}
\vspace{-15pt}
    \includegraphics[width=\linewidth]{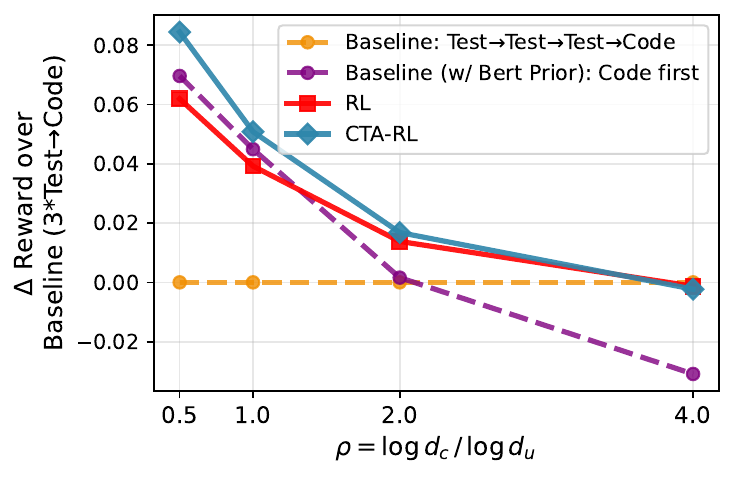}
    \caption{Average reward under varying costs for our approaches. Static strategies (test-first or code-first) achieve high reward only in limited regimes, whereas \textsc{CTA-RL} with estimated priors is consistently on the Pareto frontier}
    \label{fig:pareto-line}
    \vspace{-15pt}
\end{wrapfigure}



Figure \ref{fig:pareto-line} visualizes the $\Delta$ Reward against the naive baseline \textcolor{YellowOrange}{``3*Tests$\rightarrow$Code''} for each method.
While \textcolor{red}{\textsc{RL}} only shows advantage with large $\rho$,  and static policy such as \textcolor{purple}{Code first} only performs well with small $\rho$, our method \textcolor{MidnightBlue}{\textsc{CTA-RL}} stays at the Pareto frontier across $\rho$ values (significance tests in Appendix~\ref{app:sig-testing}). This suggests that conditioning training with explicit priors effectively reinforces the adaptive decision reasoning.

Even with imperfect prior estimates, CTA-RL consistently performs better than RL baseline across cost regimes. Appendix~\ref{app:prior_sensitivity} further analyzes sensitivity to prior estimator quality. Appendix~\ref{app:case-study} provides representative model traces comparing \textsc{RL} and \textsc{CTA-RL}, illustrating how CTA induces more cost-aware decisions.

\vspace{-5pt}
\section{Related Work}
\vspace{-5pt}
\paragraph{Decision making under incomplete information} 
LLMs are increasingly being applied to tasks with incomplete information, arising from underspecified user queries \citep{cole2023selectively,zhang2025modeling,zhang-choi-2025-clarify,li2025questbench,shaikh-etal-2025-navigating}, ambiguity \citep{min-etal-2020-ambigqa,choi2025language,Deng2025InteractCompES}, and partially observed environments \citep{Wong2023FromWM,lin-etal-2024-decision,dwaracherla2024efficient,Chen2025WhenGW,grand2025shoot}. To resolve uncertainty, models often need to ask clarifying questions \citep{rao-daume-iii-2018-learning,handa2024bayesian,lalai2025the}, query the environments \citep{CHARIKAR2002785, Nadimpalli2025NoPT,monea2024llms}, or engage in collaboration \citep{wu2025collabllm,chen-etal-2025-optima}. While prior work has typically focused on training or prompting strategies, we show that models can abstractly reason about the optimal solution when provided with explicit priors, which we use to induce such reasoning.

\vspace{-5pt}
\paragraph{Agents in cost-aware deployment}
LLM-based agents are increasingly deployed in real-world settings that require multi-step reasoning and tool use, including interactive coding \citep{NEURIPS2024_d5c56ec4,zhou2025creditbudgeted,wang-etal-2025-exploracoder,jain2025multiturn}, planning \citep{zhou2024archer,liu2025costbench}, question answering \citep{yao2023react,eisenstein2025dont}, and scientific research \citep{NEURIPS2023_ef0164c1,GXChen2025LanguageAM, Khan2025OneLT,Abaskohi2025DRBenchAR,agarwal2025open}. 
While tool use expands agents’ capabilities and reliability, interacting with external environments often incurs latency \citep{guan2025dynamic}, resources \citep{Damani2024LearningHH, liu2025budget}, and overhead that can negatively affect user experience \citep{DBLP-ElfleetC24,herlihy2024overcoming}.
In response, several lines of work have emerged to study agent behavior under explicit cost constraints. 
\citet{liu2025costbench} introduce a cost-centric benchmark for evaluating agents’ tool-planning abilities. 
\citet{wang2025acting,gul2025pay,wang2025self,Lin2025AdaSearchBP} study how models can reduce unnecessary retrieval or tool use while maintaining answer quality, for example through abstention, selective search, or efficiency-oriented action policies.
\citet{berant2025learning} train steerable clarification policies that adapt to cost coefficients.
An underexplored aspect of efficient exploration is the joint treatment of uncertainty priors and cost constraints, which together determine Pareto-optimal decisions. We propose a unified framework for interactive agentic tasks and show that calibrated priors are key to inducing appropriate decision-making in LLM agents.

\vspace{-5pt}
\section{Conclusion}
\vspace{-5pt}

This paper presents a method for having LLMs balance uncertainty-cost tradeoffs in their environment interaction. By presenting an LLM with priors over unobserved features of the environment, the LLM can successfully reason about Pareto-optimal behavior and navigate action costs effectively.  This work illustrates new ways of inducing agents to think optimally, and suggests that meta-level information (priors about capabilities) may have a role to play in shaping agent policies.

\begin{ack}
    Thanks to Xi Ye for comments on a draft of this work. This work was supported by the NSF under Cooperative Agreement 2421782 and the Simons Foundation grant MPS-AI-00010515 awarded to the NSF-Simons AI Institute for Cosmic Origins — CosmicAI, \url{https://www.cosmicai.org/}. This was was also partially supported by NSF CAREER Award IIS-2145280, NSF grant IIS-2433071, by the Sloan Foundation, and by grants from Amazon and Open Philanthropy. This research has been supported by computing support on the Vista GPU Cluster through the Center for Generative AI (CGAI) and the Texas Advanced Computing Center (TACC) at the University of Texas at Austin, through the Torch cluster at NYU, and through a compute grant from NVIDIA.
\end{ack}

\bibliography{neurips_2026}
\bibliographystyle{plainnat}






\newpage
\appendix





\section{Limitations}
\label{app:limitations}

Our experiments focus on controlled \textsc{QA} and \textsc{FileReading} settings that isolate cost-aware exploration, but they are simpler than open-ended agent deployments with richer action spaces, longer horizons, and noisier observations. 
We use these settings to enable reliable evaluation of decision quality, while extending CTA to more complex environments remains a promising direction for future work.

CTA also relies on the availability of useful prior estimates. 
Our sensitivity analysis shows that CTA is robust to moderate estimation error, but performance can degrade when priors are highly miscalibrated or confidently wrong. We demonstrate several ways to obtain priors without ground-truth access, including from model confidence and historical task data, but designing reliable prior estimators for broader domains remains future work.

Finally, CTA does not guarantee optimal behavior in arbitrary environments. 
The model must still correctly interpret the provided priors and reason about their implications for action selection. 
In more complex scenarios, optimal action selection may require stronger reasoning capabilities of the base model to account for posterior updates, multi-step value of information, and dependencies among multiple latent variables.

\section{Impact Statement}
\label{app:impact}
This paper presents a method for broadly improving the cost-benefit tradeoffs of LLM agents. Although this is not yet integrated into production agent systems, we envision that this approach or one derived from it could be, which would ideally lead to cost savings and increased efficiency. We do not foresee specific drawbacks of this approach relative to other advancements in LLMs, agents, and machine learning more broadly. Potential broad drawbacks are inherited from the drawbacks of capability advancements for LLMs and agents more broadly, such as enabling the further propagation of AI technology in society.

\section{Sensitivity to Prior Quality}
\label{app:prior_sensitivity}
We evaluate how performance varies with prior quality in the \textsc{FileReading} task by systematically degrading the prior estimator used by \textsc{CTA-RL}.
In addition to the learned BERT-based prior, we construct three alternative priors with progressively worse calibration:
\begin{itemize}
    \item \textbf{Noisy}: Derived from the BERT prior by randomly selecting $40\%$ of predictions and swapping the probabilities assigned to the correct and incorrect options.
    \item \textbf{Uniform}: A uniform distribution over all candidate options, representing the absence of informative prior knowledge.
    \item \textbf{Adversarial}: Constructed by swapping the probabilities assigned to correct and incorrect options and inflating the incorrect option to $95\%$ confidence.
\end{itemize}

\begin{table}[t]
\centering
\small
\caption{Sensitivity to prior quality in the \textsc{FileReading} task. Performance degrades smoothly as priors become less accurate, without a sharp failure transition.}

\begin{tabular}{lcccccccc}
\toprule
\textbf{Prior} & \textbf{Prior Acc} & \textbf{Avg Reward} & \textbf{Task Acc} & \#\textbf{Code} & \textbf{$R_{\rho=0.5}$} & \textbf{$R_{\rho=1.0}$} & \textbf{$R_{\rho=2.0}$} & \textbf{$R_{\rho=4.0}$} \\
\midrule
BERT        & 0.67 & 0.269 & 99.1\% & 1.46 & 0.382 & 0.315 & 0.230 & 0.147 \\
Noisy       & 0.52 & 0.251 & 96.3\% & 1.71 & 0.365 & 0.295 & 0.206 & 0.136 \\
Uniform     & 0.46 & 0.251 & 96.8\% & 1.53 & 0.349 & 0.300 & 0.219 & 0.134 \\
Adversarial & 0.33 & 0.203 & 94.4\% & 2.26 & 0.299 & 0.240 & 0.173 & 0.099 \\
\bottomrule
\end{tabular}
\label{tab:prior_sensitivity}
\end{table}

Table~\ref{tab:prior_sensitivity} reports performance across different prior qualities. As prior quality decreases, both reward and task accuracy degrade gradually. Importantly, the degradation is smooth rather than exhibiting a sharp failure transition, indicating robustness to imperfect priors.
This suggests that approximate or imperfectly estimated priors are sufficient to yield meaningful gains (even the BERT prior achieves only $67\%$ accuracy), indicating that our approach is inherently error-tolerant and applicable in complex real-world settings where accurate priors are difficult to obtain due to noisy or high-dimensional features. At the same time, the adversarial setting highlights that being confidently wrong is substantially more harmful than being uncertain, leading to the largest drop in performance. 






\section{Significance Testing for Pareto-Optimality in \textsc{FileReading}}
\label{app:sig-testing}
We evaluate whether the gains of \textsc{CTA-RL} are statistically reliable using paired bootstrap tests on per-task rewards.
For each cost setting $\rho \in \{0.5, 1.0, 2.0, 4.0\}$, we pair the test examples across methods by task ID.
We compute per-task reward differences between \textsc{CTA-RL} and each baseline, and use $10{,}000$ bootstrap resamples to test whether the mean difference is significantly different from zero.
Table~\ref{tab:pareto-significance} reports the observed mean reward gain $\Delta$.

\begin{table}[t]
\centering
\small
\caption{
Paired bootstrap tests for \textsc{CTA-RL} on \textsc{FileReading}.
$\Delta$ denotes the mean per-task reward gain of \textsc{CTA-RL} over each baseline.
Tests use $n=300$ paired tasks per $\rho$ and $B=10{,}000$ bootstrap resamples.
}
\label{tab:pareto-significance}
\begin{tabular}{c c c c}
\toprule
$\rho$ & vs. Test first (B1) & vs. Code first (B2) & vs. \textsc{RL} \\
\midrule
$0.5$ & $+0.0845^{**}$ & $+0.0149^{=}$ & $+0.0224^{**}$ \\
$1.0$ & $+0.0508^{**}$ & $+0.0059^{=}$ & $+0.0117^{*}$ \\
$2.0$ & $+0.0169^{**}$ & $+0.0153^{=}$ & $+0.0030^{=}$ \\
$4.0$ & $-0.0021^{=}$ & $+0.0287^{**}$ & $-0.0010^{=}$ \\
\bottomrule
\end{tabular}

\vspace{0.3em}
\small
$^{*}p<0.01$, $^{**}p<0.001$, and $^{=}$ denotes no significant difference.
\end{table}

Overall, \textsc{CTA-RL} achieves significant gains over different baselines in several cost regimes and is never statistically worse than any baseline. 
These results support the conclusion that \textsc{CTA-RL} remains on or near the Pareto frontier across cost regimes.

\textsc{CTA-RL} significantly outperforms the no-prior test-first baseline at $\rho \in \{0.5,1.0,2.0\}$ and is statistically indistinguishable from it at $\rho=4.0$.
Against the prior-aware code-first baseline, \textsc{CTA-RL} is statistically indistinguishable at lower $\rho$, where code-first already benefits from prior information, but significantly outperforms it at $\rho=4.0$, where code execution becomes relatively expensive.
Compared to the no-prior \textsc{RL} agent, \textsc{CTA-RL} achieves significant gains at $\rho=0.5$ and $\rho=1.0$ and is statistically indistinguishable at higher $\rho$.
Overall, these results support the conclusion that \textsc{CTA-RL} remains on or near the Pareto frontier across cost regimes.

\section{Qualitative Trace Analysis of Pandora's Box Problem}
\label{app:pandora-trace-analysis}
We present representative interaction traces from three settings: \textsc{CTA-Prompted-NonThink}, \textsc{Prompted}, and \textsc{CTA-Prompted}. In \textsc{CTA-Prompted-NonThink} (with thinking mode disabled; Figure \ref{fig:case-study-pandora-cta-prompted-nt}), the model does not appear to compare the expected value of additional information against the exploration cost. As a result, it tends to verify all options before committing, regardless of the prior distribution, leading to unnecessary exploration. In \textsc{Prompted} (Figure \ref{fig:case-study-pandora-prompted}), the model does not have access to the prior probabilities. Lacking calibrated uncertainty information, it effectively operates under an implicit uniform prior and consequently follows a suboptimal verification strategy.

In contrast, in \textsc{CTA-Prompted} (Figure \ref{fig:case-study-pandora-cta-prompted}), the model is provided with prior probabilities and has thinking mode enabled. In this setting, it explicitly reasons about the trade-off between expected reward and exploration cost by comparing the value of immediate commitment with the discounted value of further verification. The resulting behavior aligns with the oracle policy.

These qualitative examples illustrate how explicit prior information induces the correct reasoning of the model over value of additional information against action cost and making the optimal decision accordingly.






\begin{figure*}
\begin{tcolorbox}[
    title=\textbf{Case Study: Pandora's Box Problem, \textsc{CTA-Prompted-NonThink}}, 
    fontupper=\footnotesize, 
    fonttitle=\large\bfseries,
    sharp corners=south,
    coltitle=white,
]
\textbf{Problem Parameters}\\
- Bag Labels: A, B, C\\
- Prior Probabilities: A: 0.04, B: 0.68, C: 0.28\\
- Discount Factor r: 0.2\\

\textbf{Interaction Trace}

\textit{Turn 0}\\
\textbf{User: } \{Instruction instantiated with the problem parameters\}\\
\textbf{Assistant:} VERIFY A\\
\textit{Turn 1}\\
\textbf{User: } The verification result is: NO, A is incorrect. Given this, please provide your next action.\\
\textbf{Assistant:} VERIFY B\\
\textit{Turn 2}\\
\textbf{User: } The verification result is: NO, B is incorrect. Given this, please provide your next action.\\
\textbf{Assistant:} GUESS C\\

\end{tcolorbox}
\caption{Example interaction trace on a 3-bag Pandora’s Box instance with priors (0.04, 0.68, 0.28) and discount factor $\gamma=0.2$ with thinking mode \textbf{disabled}. In this setting, the model explores all bags before committing and follows a suboptimal verification order, rather than prioritizing the highest-probability option.}

\label{fig:case-study-pandora-cta-prompted-nt}
\end{figure*}

\begin{figure*}
\begin{tcolorbox}[
    title=\textbf{Case Study: Pandora's Box Problem, \textsc{Prompted}}, 
    fontupper=\footnotesize, 
    fonttitle=\large\bfseries,
    sharp corners=south,
    coltitle=white,
]
\textbf{Problem Parameters}\\
- Bag Labels: A, B, C\\
- Prior Probabilities: A: 0.04, B: 0.68, C: 0.28\\
- Discount Factor r: 0.2\\

\textbf{Interaction Trace}

\textit{Turn 0}\\
\textbf{User: } \{Instruction instantiated with the problem parameters, \textbf{but the prior probabilities are not provided to the model}\}\\
\textbf{Assistant:} 
\texttt{\textless think\textgreater} Okay, let's see. I need to figure out the best action here. So, there are three bags: A, B, and C. Only one of them has the prize worth 1. The discount factor is 0.2. At each step, I can either verify a bag to get information (which takes a timestep) or guess a bag to end the game and get the reward, which is discounted by $r^t$ where t is the timestep.

Starting at t=0. If I guess now, the reward would be $1 * 0.2^0 = 1$. But if I guess wrong, I get nothing. Since there's a $1/3$ chance of being correct, the expected reward for guessing at t=0 is $(1/3)*1 + (2/3)*0 = 1/3 \approx 0.333$.

Alternatively, if I verify a bag. Let's say I verify bag A. It takes one timestep (t=0 to t=1). If the verification says "YES", then I know A has the prize. Then, I can guess A at t=1, getting $1 * 0.2^1 = 0.2$. If the verification says "NO", then the prize is in B or C. Then, I have two bags left. But I need to think about the next steps.

Wait, but verifying takes a timestep. So, if I verify at t=0, then the next step is t=1. Let's think about the expected value of verifying. Suppose I verify A at t=0. Then, the result is YES with 1/3, NO with 2/3.

If YES: Then at t=1, I can guess A and get 0.2. If NO: Then the prize is in B or C. Now, at t=1, I have two bags left. If I guess at t=1, the expected reward would be $(1/2)*0.2^1 + (1/2)*0 = 0.1$. Alternatively, I can verify another bag. Let's say I verify B at t=1. Then, if it's YES, I can guess B at t=2 (reward $0.2^2=0.04$), else verify C at t=2, etc. But this might take longer.

Alternatively, maybe it's better to guess at t=1 if I have two bags left. The expected value would be 0.1 as above. But if I verify, maybe I can get more information. Let me calculate the expected value of verifying A at t=0.

So, if I verify A at t=0, then at t=1, if it's YES (prob 1/3), I get 0.2. If NO (prob 2/3), then I have two bags left. Then, the expected value from that point is the maximum between guessing or verifying again. Let's assume that the optimal strategy is to guess when there are two bags left. Then, the expected value would be 0.1 as before. So the total expected value of verifying A at t=0 is $(1/3)*0.2 + (2/3)*0.1 = 0.2/3 + 0.2/3 = 0.4/3 \approx 0.133$. Which is less than the 0.333 from guessing at t=0. So, it seems better to guess at t=0.

Wait, but maybe I'm missing something. Because if I verify A at t=0, and it's NO, then at t=1, I can verify another bag. Let's say I verify B at t=1. Then, if it's YES (prob 1/2), then I can guess B at t=2 (reward $0.2^2=0.04$). If NO (prob 1/2), then the prize is in C, so I can guess C at t=2, getting 0.04. So the expected value here would be 0.04. So, the total expected value if I verify A at t=0, then B at t=1, would be $(1/3)*0.2 + (2/3)*[ (1/2)*0.04 + (1/2)*0.04 ] = (1/3)*0.2 + (2/3)*0.04 = 0.0667 + 0.0267 \approx 0.0933$. Which is worse than the 0.333 from guessing at t=0. So, in that case, guessing at t=0 is better.

Alternatively, maybe there's a better strategy. Let me think again. Suppose I verify one bag, and if it's no, then verify another. But with discounting, the later rewards are smaller. So, perhaps the optimal strategy is to guess immediately. Because the expected reward from guessing at t=0 is 1/3, which is higher than the expected value of verifying and then proceeding. So, maybe the best action is to GUESS at t=0.

But wait, what if I verify one bag and then guess? Let's say I verify A at t=0. If it's YES, then I can guess at t=1 (0.2). If it's NO, then I have two bags left. Then, if I guess at t=1, the expected reward is 0.1. So the total expected value is $(1/3)*0.2 + (2/3)*0.1 = 0.2/3 + 0.2/3 = 0.4/3 \approx 0.133$, which is less than 0.333. So, indeed, guessing at t=0 is better.

Alternatively, if I verify two bags, but that would take more steps. For example, verify A at t=0, then B at t=1. If both are NO, then the prize is in C. Then at t=2, guess C and get $0.2^2 = 0.04$. So the expected value would be (probability that A is NO and B is NO) * 0.04. The probability that A is NO is $2/3$, then B is NO is $1/2$, so $2/3 * 1/2 = 1/3$. So expected value is $1/3 * 0.04 = 0.0133$. Which is worse than guessing at t=0. So, that's even worse.

So, the conclusion is that guessing at t=0 gives a higher expected value than any strategy involving verification. Therefore, the optimal action is to GUESS one of the bags. But since all bags are symmetric, it doesn't matter which one. So, the answer should be GUESS A, B, or C. But the user probably expects me to choose one. Let me check if there's any other consideration. For example, maybe the discount factor is 0.2, which is quite low. So, waiting to get more information might not be worth it. So, the best action is to GUESS now.
\texttt{\textless/think\textgreater}

GUESS A

\end{tcolorbox}
\caption{Example interaction trace on a 3-bag Pandora’s Box instance with priors (0.04, 0.68, 0.28) and discount factor $\gamma=0.2$, where the model is not given access to the prior probabilities. In this setting, the model implicitly treats the bags as equally likely and follows a suboptimal strategy that deviates from the optimal policy.}
\label{fig:case-study-pandora-prompted}
\end{figure*}

\begin{figure*}
\begin{tcolorbox}[
    title=\textbf{Case Study: Pandora's Box Problem, \textsc{CTA-Prompted}}, 
    fontupper=\footnotesize, 
    fonttitle=\large\bfseries,
    sharp corners=south,
    coltitle=white,
]
\textbf{Problem Parameters}\\
- Bag Labels: A, B, C\\
- Prior Probabilities: A: 0.04, B: 0.68, C: 0.28\\
- Discount Factor r: 0.2\\
\\
\textbf{Interaction Trace}\\
\textit{Turn 0}\\
\textbf{User: } \{Instruction instantiated with the problem parameters\}\\
\textbf{Assistant:}\\
\texttt{\textless think\textgreater}
Okay, let's see. I need to figure out whether to verify or guess in this scenario. The problem has three bags: A, B, and C with prior probabilities 0.04, 0.68, and 0.28 respectively. The discount factor is 0.2. 

First, I should remember that each VERIFY action costs a timestep, and GUESS ends the game immediately. The reward for a correct guess is discounted by $r^t$, where t is the timestep. So, if I guess at $t=0$, the discount is 1 (since $0.2^0 = 1$). But if I verify, I spend a timestep and then maybe have more information to make a better decision.

The key here is to balance between getting more information (which might lead to a higher reward) and the risk of waiting too long, which could reduce the discounted reward. 

Let me think about the expected value of guessing immediately versus verifying. 

If I guess at $t=0$, the expected reward would be the sum of the probabilities of each bag multiplied by the discount factor. Wait, no. Wait, the reward is 1 if the guessed bag has the prize, 0 otherwise. So the expected reward for guessing at t=0 would be the probability of the chosen bag having the prize. \hl{But since I can choose any bag to guess, I should pick the one with the highest prior probability. }

Looking at the priors: B is 0.68, C is 0.28, A is 0.04. So if I guess B, the expected reward is 0.68 * 1 (since if it's correct, I get 1, else 0). But since the discount is $0.2^0 = 1$, the expected reward is 0.68. 

Alternatively, if I verify a bag, I can get more information. Let's say I verify B first. If I verify B, then I spend a timestep (t=1). Then, depending on the result, I can decide whether to guess or verify again. 

...

But the problem is that the discount factor is 0.2. So, each time you wait, the reward is discounted. So, the longer you wait, the less valuable the reward becomes. 

So, the question is \hl{whether the information gained from verifying is worth the cost of waiting}. 

Let me think about the expected value of verifying versus guessing. 

\hl{If I guess at t=0}, the expected reward is the maximum prior probability, which is \hl{0.68}. 

\hl{If I verify a bag, say B, then I spend one time step (t=1). Let's calculate the expected reward if I verify B. }

\hl{After verifying B, there are two possibilities: }

1. The result is YES (probability 0.68). Then, I know B has the prize. So, I can guess B at t=1, which would give me a reward of $1 * 0.2^1 = 0.2$. 

2. The result is NO (probability $1 - 0.68 = 0.32$). Then, B does not have the prize. Now, I have to decide whether to guess or verify another bag. 

But if I get NO on B, then the remaining bags are A and C. Their probabilities would be updated based on the fact that B is not the correct one. Wait, but the problem says that the bags are independent. Wait, actually, the problem states that the prior probabilities are for each bag. So, if B is not the correct one, then the prize is in A or C. However, the original priors are A:0.04, B:0.68, C:0.28. But if we know that B is not the correct one, then the total probability of the prize being in A or C is $0.04 + 0.28 = 0.32$. \hl{So, the posterior probabilities would be A: $0.04 / 0.32 = 0.125$, and C: $0.28 / 0.32 = 0.875$.} 

So, in this case, if we get a NO on B, then the optimal choice would be to guess C, which has a 0.875 chance of being correct. Then, the reward would be $0.875 * 0.2^1$ (since t=1). 

But wait, after verifying B at t=0, we spend a timestep (t=1), and then we can guess. So, if we get NO on B, then we can guess C at t=1, which gives us $0.875 * 0.2 = 0.175$. 

Alternatively, if we get YES on B, then we can guess B at t=1, which gives $1 * 0.2 = 0.2$. 

\hl{So, the expected reward of verifying B is: 

$0.68 * 0.2 + 0.32 * 0.175 = 0.136 + 0.056 = 0.192$. }

Compare that to the expected reward of guessing immediately (0.68). 

\hl{So, 0.68 is higher than 0.192. Therefore, verifying B is worse than guessing immediately. }

\texttt{\textless/think\textgreater}

Action: GUESS B

\end{tcolorbox}
\caption{Example model reasoning trace on a 3-bag Pandora’s Box instance with priors (0.04, 0.68, 0.28) and discount factor $\gamma=0.2$. The model explicitly compares the expected value of immediate guessing versus verification and then chooses to guess $B$ immediately, which is the optimal strategy in this case. Key reasoning steps, including the explicit comparison between action value and exploration cost, are highlighted in yellow. This example illustrates that when priors are provided explicitly, the model can reason about uncertainty and exploration cost to select an optimal strategy.}

\label{fig:case-study-pandora-cta-prompted}
\end{figure*}

\section{Experiment Details for \textsc{QA}}
For \textsc{QA}, we evaluate on the \textsc{PopQA} dataset. We build the retriever based on Contriever \citep{izacard2022unsupervised,bajaj2016ms}. The retriever quality $p_\text{ret}$, defined as the probability that the model can answer a question correctly when conditioned on the document retrieved from the retriever, is estimated on a validation set and provided to the agent as a part of the input $\mathbf{x}$ at inference time. Note that $p_\text{ret}$ depends on the retriever and the agent being used, but not on individual questions.

\section{Dataset Construction Details for Task: Coding with Selective Testing}
\label{app:code-data-generation}
This appendix describes the oracle filename-to-format model and the procedure used to generate the coding task dataset \textsc{FileReading}.

Each task consists of a structured filename, a CSV file generated under a latent formatting configuration, and an associated query.
Filenames combine a small set of indicative tokens (e.g., \texttt{\_eu}, \texttt{\_tab}, \texttt{\_sas}, \texttt{\_cn}) with additional irrelevant strings, inducing a prior over possible parsing configurations.

To capture scenarios in which filename cues provide informative signals about file formats, we construct a synthetic dataset of CSV-based question-answering tasks with structured filenames. Each filename is generated by combining a small set of indicative tokens (e.g., \texttt{\_eu}, \texttt{\_tab}, \texttt{\_sas}, \texttt{\_cn}) with additional irrelevant strings. 

We define an oracle filename-to-format model
\[
\mathcal M_{\text{oracle}} : n \;\mapsto\; p(\mathbf z \mid n),
\]
which maps a filename \(n\) to a prior distribution over formatting configurations
\(\mathbf z=(z_d,z_q,z_s)\).
For example, filename tokens like \texttt{\_tsv} substantially increase the probability of a tab delimiter relative to a default comma delimiter.

We then sample the true formatting configuration \(\mathbf z^* \sim p(\mathbf z \mid n)\) and generate the corresponding CSV content and task query.

The dataset is constructed so that the correct answer is obtainable only when the file is parsed with the correct configuration; incorrect formatting assumptions lead to parsing failures or misaligned columns. 
Filenames are represented by four binary features, yielding \(2^4=16\) distinct filename feature configurations and corresponding prior distributions over formatting attributes.
We generate 2{,}000 task instances, each consisting of a filename, a CSV file generated under a sampled formatting configuration, and an associated query.
The dataset is split into 1{,}400 training examples, 300 validation examples, and 300 test examples. 
Details of the oracle filename-to-format model, feature templates.

\paragraph{Latent Formatting Variables}
Each task instance is associated with a latent formatting configuration
\[
\mathbf z = (z_d, z_q, z_s),
\]
where
\(z_d \in \{\,\texttt{,},\texttt{;},\texttt{\textbackslash t}\,\}\) denotes the delimiter,
\(z_q \in \{\,\texttt{"},\texttt{'}\,\}\) the quote character, and
\(z_s \in \{0,1\}\) the number of header rows to skip.
The correct answer can be obtained if and only if the file is parsed using the fully correct configuration \(\mathbf z^*\).

\paragraph{Filename Features}
We extract four binary features from each filename \(n\), each indicating the presence or absence of a specific substring:
\texttt{has\_eu}, \texttt{has\_tsv}, \texttt{has\_sas}, and \texttt{has\_cn}.
Each feature is either on or off, resulting in \(2^4 = 16\) possible filename feature configurations.
Each configuration corresponds to a distinct prior distribution over formatting attributes induced by the oracle model.

\paragraph{Oracle Filename-to-Format Model}
We define an oracle filename-to-format model
\[
\mathcal M_{\text{oracle}} : n \mapsto p(\mathbf z \mid n),
\]
which maps a filename \(n\) to a prior distribution over formatting configurations
\(\mathbf z=(z_d,z_q,z_s)\).
For each of the 16 possible filename feature configurations, the model induces a corresponding prior over formatting attributes.
The prior factorizes as
\[
p(\mathbf z \mid n)
= [p(z_d \mid n),p(z_q \mid n), p(z_s \mid n)],
\]
where each factor is parameterized as a log-linear model over the filename features.

\paragraph{Sampling and File Generation}
For each filename \(n\), we sample a formatting configuration
\(\mathbf z \sim p(\mathbf z \mid n)\)
from the oracle model $\mathcal M_{\text{oracle}}$.
We then generate a CSV file whose content conforms to \(\mathbf z\).
The data are constructed such that incorrect parsing—due to an incorrect delimiter, quote character, or number of skipped rows—either produces malformed outputs or prevents access to the correct answer.

\paragraph{Task Instances and Splits}
Each task instance consists of a filename \(n\), a generated CSV file, and a query requiring the agent to compute an answer from the file.
We generate 2{,}000 task instances in total, split into 1{,}400 training examples, 300 validation examples, and 300 test examples.

\section{Oracle Strategy for Pandora's Box Problem}\label{app:oracle-proof}
Algorithm \ref{alg:pandora-k3} presents the optimal policy for the Pandora's Box problem. In this section, we prove its optimality.

We begin by characterizing the structure of an optimal policy. 
At any state, let $S$ denote the remaining set of boxes and let $q_i(S)$ denote the posterior probability that box $i \in S$ contains the prize. 
An optimal policy only needs to consider the box with maximum posterior probability at each step.

First, if the agent commits, the expected reward equals the posterior success probability of the chosen box. Hence, committing to any box other than a maximum-posterior box is suboptimal. 
Second, verification is beneficial only insofar as it increases the probability of early termination before further discounting. Verifying a higher-posterior box increases the chance of immediate success and therefore weakly dominates verifying a lower-posterior box. Consequently, it suffices to consider the box
$i^\star \in \arg\max_{i \in S} q_i(S)$
at each decision step.

It remains to determine whether the agent should commit to $i^\star$ immediately or verify it first. 
Let $q = q_{i^\star}(S) = \max_{i \in S} q_i(S)$ be the posterior probability of the most likely box $i^\star$. 

If the agent commits immediately, the expected value is
\[
V_{\mathrm{guess}}(S) = q.
\]

If it verifies $i^\star$, then with probability $q$ verification succeeds and yields reward $1$, and with probability $1-q$ it fails and the problem reduces to the smaller set $S \setminus \{i^\star\}$. Since verification incurs one multiplicative discount factor $\gamma$, the expected value of verifying is
\[
V_{\mathrm{verify}}(S)
= \gamma \Big( q + (1-q)\, V(S \setminus \{i^\star\}) \Big).
\]

Therefore, optimality implies the Bellman recursion
\[
V(S)
= \max \big\{ V_{\mathrm{guess}}(S),\; V_{\mathrm{verify}}(S) \big\},
\]
with base case $V(\{i\}) = 1$. This recursion is exactly implemented by Algorithm~\ref{alg:pandora-k3}.







\begin{algorithm}[t]
\caption{Oracle policy for Pandora's box with $n$ boxes}
\label{alg:pandora-k3}

\begin{algorithmic}
\Function{Solve}{$S$}
    \If{$|S| = 1$}
        \State \Return $(1, \textsc{Commit}(\text{only element of } S))$ 
        \Comment{Base case: Final box selection}
    \EndIf
    \Statex
    
    \State $W \gets \sum_{j \in S} p_j$
    \State $i^\star \gets \arg\max_{i \in S} p_i$ 
    \Comment{Select candidate with highest success posterior}
    \State $q \gets p_{i^\star} / W$
    \Statex
    
    \State $V_{\mathrm{guess}} \gets q$ \Comment{Expected value if committing to box $i^*$ now}
    \State $(V_{\mathrm{fail}}, \_) \gets \Call{Solve}{S \setminus \{i^\star\}}$ \Comment{Recurse to find value if box $i^*$ is empty}
    \State $V_{\mathrm{verify}} \gets \gamma \big(q + (1-q) \cdot V_{\mathrm{fail}}\big)$ \Comment{The value of verifying $i^*$ first}
    \Statex
    
    \If{$V_{\mathrm{guess}} \ge V_{\mathrm{verify}}$}
        \State \Return $(V_{\mathrm{guess}}, \textsc{Commit}(i^\star))$
    \Else
        \State \Return $(V_{\mathrm{verify}}, \textsc{Verify}(i^\star))$
    \EndIf
\EndFunction
\end{algorithmic}

\end{algorithm}



\section{GRPO Training Details}
\label{app:grpo-training-details}
We use the GRPO implementation of the veRL library~\citep{sheng2024hybridflow, zheng2025easyr1} for the RL training.
The model is trained with a group size of $n=8$ and a KL coefficient $\beta=0.01$.
During rollout, tokens are sampled with a temperature of $1.0$.
We use a learning rate of $1 \times 10^{-6}$ and train with a global batch size of $16$.
We use a micro-batch size of $2$ for model updates and a micro-batch size of $4$ for experience.
We conduct GRPO training on Qwen3-8B using a single node with $8$ NVIDIA A100 GPUs.

\section{Prompt Templates}
\label{sec:prompt-all}
\subsection{Prompts for \textsc{Pandora}}
Prompt templates used for \textsc{Pandora} are shown in Figure~\ref{fig:prompt-pandora}.
\label{sec:prompt-pandora}
\begin{figure*}
\begin{tcolorbox}[
    title=\textbf{Prompts: Pandora's Box}, 
    fontupper=\footnotesize, 
    fonttitle=\large\bfseries,
    sharp corners=south,
    coltitle=white,
]

{\textbf{SYSTEM}} \\
\texttt{You are a rational agent tasked with solving sequential decision-making problems under uncertainty. You are given a set of options (bags) with prior probabilities of containing a prize with value 1. You can either VERIFY an option to get information (YES/NO) or GUESS an option to end the game and collect the reward. }

\texttt{- Each VERIFY action consumes one timestep. \\
- The reward for a correct GUESS is discounted by a factor $r^t$, where t is the timestep when you GUESS. \\
- You must balance information gathering (VERIFY) with timely exploitation (GUESS) to maximize expected discounted reward. }

\texttt{Always respond with exactly one action token per step, using the format: \\
VERIFY <Option> or GUESS <Option>}.\\

\textbf{Instruction}
\begin{verbatim}
--- NEW GAME ---
TIMESTEP: t=0

PROBLEM PARAMETERS:
- Bag Labels: {labels_str}
- Prior Probabilities: {priors_str} //only provided to the model in CTA- settings
- Discount Factor r: {r}

Choose your action.
\end{verbatim}

\end{tcolorbox}
\caption{Prompt template for Pandora's Box setting.}
\label{fig:prompt-pandora}
\end{figure*}

\subsection{Prompts for \textsc{QA}}
Prompts used in \textsc{QA} are provided in Figure \ref{fig:prompt-popqa}.
\label{sec:prompt-qa}
\begin{figure*}
\begin{tcolorbox}[
    title=\textbf{Prompts: QA with Optional Retriever}, 
    fontupper=\footnotesize, 
    fonttitle=\large\bfseries,
    sharp corners=south,
    coltitle=white,
]

{\textbf{SYSTEM}} \\
{You are a rational agent tasked with answering factual questions under uncertainty. At each step, you can either directly answer the question or retrieve additional context before answering. 
\\
Available actions:\\
- RETRIEVE: request a related context to consult before deciding your final answer. This consumes one timestep.\\ 
- ANSWER: $<$your short factual answer$>$, provide your final answer and end the interaction.\\
\\
Each question comes with your estimated probabilities of answering correctly:\newline
- p\_no\_context: probability you can answer correctly without retrieving.\newline
- p\_with\_context: probability you can answer correctly after retrieval.
\\\\
Your goal is to maximize expected discounted reward:\newline
Reward = $r^t *$ correctness, where t is the timestep when you issue ANSWER and correctness $\in$ \{0,1\}.
\\\\
Retrieval information:\newline
If you choose RETRIEVE, your expected answer accuracy after retrieval is {p\_with\_context}.
If you choose not to retrieve, you must rely only on your current knowledge.
\\
Be deliberate — retrieving may improve accuracy but reduces reward due to time discounting. Balance speed and correctness carefully.
\\
Always respond with exactly one action token per step, using the format: \\
RETRIEVE or ANSWER: $<$short factual answer$>$.}\\

\textbf{Instruction - turn 1}\\
\texttt{--- NEW QUESTION ---\newline
TIMESTEP: t=0
\\
Question: {question}\\
Parameters:\\
- Discount factor (r): \{r\}\\
- Success probability with retrieval (p\_with\_context): \{p\_with\_context\}\\
\\
Choose your action:\\
RETRIEVE or ANSWER: <short factual answer>.}\\

\textbf{Instruction - turn 2}\\
\texttt{TIMESTEP: t=1\\
You have retrieved the following context:\\
\{context\}\\
\\
Question: \{question\}\\
Now decide whether to answer:\\
Respond with:\\
ANSWER: <short factual answer>}
\vspace{5pt}

\end{tcolorbox}
\caption{Prompt templates for \textsc{QA}.}
\label{fig:prompt-popqa}
\end{figure*}

\subsection{Prompts for \textsc{FileReading}}
Prompts used in the \textsc{FileReading} setting are provided in Figures~\ref{fig:system-prompt-csv-main}, \ref{fig:system-prompt-guidance},~\ref{fig:instruction-prompt-csv}, and \ref{fig:instruction-prompt-csv-likelihoods}.
\label{sec:prompt-code}
\begin{figure*}[t]
\begin{tcolorbox}[
  title=\textbf{System Prompt: \textsc{FileReading}},
  fontupper=\footnotesize, 
    fonttitle=\large\bfseries,
    sharp corners=south,
    coltitle=white,
]
You are an agent tasked with writing code to fulfill an instruction about a CSV file (e.g., answering a question using its contents). Your goal is to produce a correct answer while efficiently using available resources, as measured by discounted reward.

\medskip
The exact CSV formatting may not be fully known. In practice, you can either proceed using reasonable default assumptions about the format, or run unit tests to verify specific formatting details you are unsure about before committing to a final answer.

\medskip
\textbf{Allowed actions (choose exactly ONE per turn):}

\medskip
\textbf{1) UNIT\_TESTS}\\
Run unit tests to debug CSV formatting assumptions. Unit test outputs are perfectly reliable.\\
Available unit tests:
\begin{itemize}[leftmargin=14pt, itemsep=1pt, topsep=2pt]
  \item \texttt{test\_delimiter(path)} $\rightarrow$ \{`,', `;', `\textbackslash t'\}
  \item \texttt{test\_quotechar(path)} $\rightarrow$ \{`"', ``\ '\ ''\}
  \item \texttt{test\_skiprows(path)} $\rightarrow$ \{0, 1\}
\end{itemize}

\noindent Format (NO code fences):\\
\texttt{UNIT\_TESTS: test\_delimiter("file.csv"), test\_quotechar("file.csv")}

\medskip
You may include multiple unit tests in a single UNIT\_TESTS action. Each individual unit test counts toward the total number of unit tests used.

\medskip
\textbf{2) CODE}\\
Write Python code toward solving the task using your current assumptions about the CSV format.
\begin{itemize}[leftmargin=14pt, itemsep=1pt, topsep=2pt]
  \item Enclose code in \texttt{```python ... ```} 
  \item You may import pandas as pd and read the file with:\\
  \texttt{pd.read\_csv(filepath, delimiter=..., quotechar=..., skiprows=...)}
  \item Do NOT print the entire CSV.
  \item If your code computes the final result, print it to stdout so it can be read from the output.
\end{itemize}

\noindent After submission, the code will be executed and its stdout and stderr will be returned. You may use this feedback to extract the answer, debug, run additional unit tests, refine, or write additional CODE.

\medskip
\textbf{3) ANSWER}\\
Provide the final answer to the task and end the conversation.\\
Format exactly: \texttt{ANSWER: <your\_answer>}\\
The conversation ends immediately after you provide ANSWER.

\medskip
\textbf{Reward:}
\begin{itemize}[leftmargin=14pt, itemsep=1pt, topsep=2pt]
  \item Let $U$ be the total number of unit tests used.
  \item Let $C$ be the total number of CODE actions taken.
  \item Final reward $=$ correctness $\times (d_{\text{unit}})^U \times (d_{\text{code}})^C$.
  \item Discount factors represent cost multiplicatively.
  \item A smaller discount factor means a MORE expensive action.
  \item If $d_{\text{code}} = d_{\text{unit}}^k$, one CODE attempt costs about as much as $k$ UNIT\_TESTS.
\end{itemize}
\end{tcolorbox}
\caption{System prompt for \textsc{FileReading}.}
\label{fig:system-prompt-csv-main}
\end{figure*}

\begin{figure*}[t]
\begin{tcolorbox}[
  title=\textbf{System Prompt: \textsc{FileReading} (continuation)},
  fontupper=\footnotesize, 
    fonttitle=\large\bfseries,
    sharp corners=south,
    coltitle=white,
]
\textbf{General guidance:}
\begin{itemize}[leftmargin=14pt, itemsep=1pt, topsep=2pt]
  \item Start from reasonable default beliefs about the CSV format based on common conventions or provided likelihoods.
  \item Both UNIT\_TESTS and CODE are costly actions; neither should be treated as free.
  \item Use UNIT\_TESTS to reduce uncertainty when the expected benefit outweighs their cost.
  \item Use CODE to make progress toward solving the task, but recognize that failed or repeated CODE attempts are also costly.
  \item Decide when it is better to verify assumptions with UNIT\_TESTS versus attempting CODE earlier, taking into account your confidence and the relative cost of these actions.
  \item Decide rationally how much debugging and iteration is worthwhile before committing to a final ANSWER.
\end{itemize}
\end{tcolorbox}
\caption{Continuation of the system prompt for \textsc{FileReading}.}
\label{fig:system-prompt-guidance}
\end{figure*}

\begin{figure*}[t]
\begin{tcolorbox}[
  title=\textbf{Instruction Prompt Template for \textsc{FileReading} (without estimated prior)},
  fontupper=\footnotesize, 
    fonttitle=\large\bfseries,
    sharp corners=south,
    coltitle=white,
]
You are given a CSV file \texttt{\{csv\_name\}}.

\medskip
Your task: \texttt{\{task\_description\}}

\medskip
\textbf{Additional context:}
\begin{itemize}[leftmargin=14pt, itemsep=1pt, topsep=2pt]
  \item No format likelihoods are provided.
  \item Make reasonable default assumptions about the CSV format based on common conventions, unless you choose to verify them with unit tests.
\end{itemize}

\medskip
\textbf{Reward parameters:}
\begin{itemize}[leftmargin=14pt, itemsep=1pt, topsep=2pt]
  \item Unit test discount $d_{\text{unit}}$: \texttt{\{d\_unit\}}
  \item Code iteration discount $d_{\text{code}}$: \texttt{\{d\_code\}}
\end{itemize}


\medskip
\textbf{Constraints:}
\begin{itemize}[leftmargin=14pt, itemsep=1pt, topsep=2pt]
  \item You should never print all rows of the CSV or you will get zero reward.
  \item You may use UNIT\_TESTS, CODE, or ANSWER as described in the system instructions in any order; only the final ANSWER ends the conversation.
  \item Incorrect intermediate CODE does not end the episode; only the final ANSWER determines correctness.
\end{itemize}
\end{tcolorbox}
\caption{Instruction prompt template specifying the CSV task, reward parameters, and constraints provided to the agent.}
\label{fig:instruction-prompt-csv}
\end{figure*}

\begin{figure*}[t]
\begin{tcolorbox}[
  title=\textbf{Instruction Prompt Template for \textsc{FileReading} (with CTA)},
  fontupper=\footnotesize, 
    fonttitle=\large\bfseries,
    sharp corners=south,
    coltitle=white,
]
You are given a CSV file \texttt{\{csv\_name\}}.

\medskip
Your task: \texttt{\{task\_description\}}

\medskip
\textbf{Additional context:}
\begin{itemize}[leftmargin=14pt, itemsep=1pt, topsep=2pt]
  \item Estimated format likelihoods are provided below.
  \item These likelihoods reflect how likely each formatting option is in practice and can be used as default assumptions.
\end{itemize}

\medskip
\textbf{Format likelihoods:}

\noindent\texttt{\{prior\}}

\medskip
\textbf{Reward parameters:}
\begin{itemize}[leftmargin=14pt, itemsep=1pt, topsep=2pt]
  \item Unit test discount $d_{\text{unit}}$: \texttt{\{d\_unit\}}
  \item Code iteration discount $d_{\text{code}}$: \texttt{\{d\_code\}}
\end{itemize}


\medskip
\textbf{Constraints:}
\begin{itemize}[leftmargin=14pt, itemsep=1pt, topsep=2pt]
  \item You should never print all rows of the CSV or you will get zero reward.
  \item You may use UNIT\_TESTS, CODE, or ANSWER as described in the system instructions.
\end{itemize}
\end{tcolorbox}
\caption{Instruction prompt template with estimated CSV format likelihoods, enabling the agent to use probabilistic defaults when trading off unit tests, code execution, and early commitment.}
\label{fig:instruction-prompt-csv-likelihoods}
\end{figure*}

\section{Case Study: Cost-Aware Decision Traces in \textsc{FileReading} with \textsc{CTA-RL} and \textsc{RL}}
\label{app:case-study}
Figures~\ref{fig:case-study-rl-trace} and~\ref{fig:case-study-rl-ep-trace} compare representative traces under a high relative code cost setting ($\rho = 4.0$). The RL model (trained without conditioning on explicit priors) tends to default to running unit tests before attempting any code, and does not explicitly reason about the relative costs of \textsc{UNIT\_TESTS} versus \textsc{FileReading} when choosing its next action. In contrast, the CTA-RL model exhibits the intended cost-aware behavior: it reasons about both (i) uncertainty over the CSV format and (ii) the relative cost of unit tests and code execution, and uses these factors to decide whether verification is worth performing before committing to a code attempt.

\begin{figure*}[t]
\begin{tcolorbox}[
  title=\textbf{Case study: Example thinking trace of RL model (without estimated priors) ($\rho = 4.0$)},
  fontupper=\footnotesize, 
    fonttitle=\large\bfseries,
    sharp corners=south,
    coltitle=white,
]

\textbf{Content:}

\medskip
\texttt{<think>}\\
Okay, let's tackle this problem. The user has a CSV file named `race\_tsv\_sas.tsv` and they want the minimum salary value, excluding any None entries. The file's name ends with .tsv, which usually stands for Tab-Separated Values. So, the delimiter is likely a tab character. But I should confirm that.

First, I need to figure out the CSV format. Since it's a TSV, the delimiter is probably a tab. But maybe the file uses a different delimiter. To be safe, I should run a unit test to check the delimiter. The \texttt{test\_delimiter} function can help here. If the delimiter is a tab, then the test should return `\textbackslash t`.

Next, the quote character. In TSVs, sometimes fields are enclosed in quotes, but it's not always the case. The default quotechar for pandas is double quotes, but maybe this file uses single quotes. Testing with \texttt{test\_quotechar} would help.

Also, sometimes CSV files have header rows that need to be skipped. The \texttt{test\_skiprows} function can check if there's a need to skip rows. But since the file is a TSV, maybe the first row is the header. However, without knowing the exact structure, it's safer to assume that the first row is the header and set \texttt{skiprows=0}. But maybe the file has some leading rows to skip.

But since the user mentioned it's a TSV, I can start by assuming the delimiter is a tab. Let me run the \texttt{test\_delimiter} first. If the delimiter is a tab, then I can proceed. If not, I'll adjust accordingly.

Once the delimiter is confirmed, I can use pandas to read the CSV. Then, check the `salary` column, filter out None values, and find the minimum.

But to avoid errors, maybe I should first run the unit tests to confirm the delimiter and quotechar. Let's start with that.\\
\texttt{</think>}

\medskip
\textbf{Action:}\\
\texttt{UNIT\_TESTS: test\_delimiter("race\_tsv\_sas.tsv"), test\_quotechar("race\_tsv\_sas.tsv")}

\end{tcolorbox}
\caption{Example reasoning trace of an RL-trained model without explicit prior conditioning in the CSV exploration task. Despite operating under the same high relative code cost setting ($\rho = 4.0$), the model defaults to verification-first behavior based on surface cues (e.g., file extension) and does not explicitly reason about uncertainty or cost trade-offs, illustrating a lack of adaptive decision-making compared to the CTA-RL model.}
\label{fig:case-study-rl-trace}
\end{figure*}

\begin{figure*}[t]
\begin{tcolorbox}[
  title=\textbf{Case study: Example thinking trace of CTA-RL model ($\rho = 4.0$)},
  fontupper=\footnotesize, 
    fonttitle=\large\bfseries,
    sharp corners=south,
    coltitle=white,
]

\textbf{Content:}

\medskip
\texttt{<think>}\\
Okay, let's tackle this problem. The task is to compute the average of the `salary` column from the CSV file `must\_eu.csv`, excluding any None values. The challenge is figuring out the correct CSV format parameters to read the file properly.

First, I need to determine the delimiter, quotechar, and skiprows parameters. The format likelihoods are given, so I should start with the most probable options. The delimiter is most likely to be `;` with a probability of $\sim$0.85. The quotechar is probably `"` with 0.589, and skiprows is 0 with 0.607. But I'm not 100\% sure, so maybe I should run some unit tests to confirm.

But wait, the reward parameters are important. Each unit test costs 0.77 per use, and code attempts cost 0.3515 per use. Since $d_{\text{code}} = d_{\text{unit}}^4$, one code attempt is equivalent to 4 unit tests. So, if I'm uncertain, it's cheaper to run unit tests first. However, if I can make an educated guess, maybe I can proceed with code.

Let me think: the delimiter is most likely `;`, so maybe I can try reading the file with that. But if the actual delimiter is different, the code will fail. Alternatively, running a unit test for the delimiter would be better. Let's check the delimiter first.

So, I'll run UNIT\_TESTS to check the delimiter. The \texttt{test\_delimiter} function will tell me if it's comma, semicolon, or tab. Since the likelihood is high for semicolon, but I need to confirm. Let's do that.

Once I know the delimiter, I can check the quotechar. The quotechar is either `"` or `'`, with higher likelihood for `"` (0.589). But again, maybe a unit test is better. However, if I proceed with code, maybe I can handle the possible quotechar. But if the code fails, I'll have to adjust.

Alternatively, maybe I can proceed with the most likely parameters and see. Let's try that. Let's assume delimiter is `;`, quotechar is `"`, and skiprows 0. Then, read the CSV with these parameters, compute the average of salary, skipping None values.

But if the actual delimiter is different, the code will fail. So, perhaps it's better to run unit tests first. Let's start with the delimiter.

So, the first action is to run UNIT\_TESTS to determine the delimiter and quotechar. Let's do that.\\
\texttt{</think>}

\medskip
\textbf{Action:}\\
\texttt{UNIT\_TESTS: test\_delimiter("must\_eu.csv"), test\_quotechar("must\_eu.csv")}

\end{tcolorbox}
\caption{Example reasoning trace of the CTA-RL model on the \textsc{FileReading} task ($\rho = 4.0$), illustrating cost-aware trade-offs between unit tests and code execution under a high relative code cost setting, while jointly reasoning about format uncertainty.}
\label{fig:case-study-rl-ep-trace}
\end{figure*}











\clearpage
\input{checklist.tex}

\end{document}

%% file: tables/symbol-mapping.tex
\centering
\small
\caption{Unified formalization of cost-aware decision problems and their instantiations across tasks. We characterize each by latent variables $z^*$, a prior belief $p$ over $\mathbf{z}$, an action space $\mathcal{A}$ consisting of exploration and commit actions, observations $\mathcal{O}$ revealed through exploration, costs $\theta$ associated with the exploration actions, and a final reward $R$ that discounts task success by incurred costs.}
\resizebox{\linewidth}{!}
{\begin{tabular}{m{2.0cm} m{2.8cm} m{2.8cm} m{1.5cm} m{2cm} m{1.1cm} m{2.3cm}}
\toprule
\textbf{Setting}
& $z^* \in \mathcal Z$
& $p$
& $\mathcal A$
& $\mathcal O$
& $\theta$
& $R$ \\
\midrule[0.75pt]
\textsc{Unified} (\S \ref{sec:unified-form})
& Latent env.~state
& Prior over relevant latent variables
& Actions
& Observations
& Cost
& Reward \\
\midrule[0.75pt]
\textsc{Pandora} (\S \ref{sec:poc})
& $z^* \in \{z_1,\ldots,z_K\}$ \newline Prize-containing box
& $\{p_k = \pi(z^* = z_k)\}$
& Verify($k$), \newline Commit($k$)
& Whether $z_k$ has prize
& $\gamma$
& $\gamma^{T}\mathbb{I}(z_k=z^*)$ \\
\midrule[0.75pt]
\textsc{QA} (\S \ref{sec:qa-form})
& $(p_\text{da}, p_\text{ret})$ \newline {Answer probabilities with and without retrieval}
& $p_{\text{da}} = \delta(z = k_{\text{da}}(\mathbf{x}))$, $p_{\text{ret}} = \delta(z = k_{\text{ret}})$
& Retrieve, \newline Answer($a$)
& Retrieved \newline context
& $\gamma$
& $\gamma^{T}\mathbb{I}(a=a^*)$ \\
\midrule[0.75pt]
\textsc{Code} (\S \ref{sec:code-form})
& $\mathbf z^*=(z_d^*,z_q^*,z_s^*)$ \newline $\in \mathcal Z_d \times \mathcal Z_q \times \mathcal Z_s$
& $p(\mathbf{z} \mid n)$ 
& \texttt{UNIT\_TEST}($f$), \newline \texttt{CODE}($d$, $q$, $s$), \newline \texttt{ANSWER($a$)}
& Format value, \newline stdout + stderr
& $(d_u, d_c)$
& $d_u^{U} d_c^{C}\mathbb{I}(a=a^*)$ \\
\bottomrule
\end{tabular}}
\label{tab:symbol-mapping}

%% file: checklist.tex
\section*{NeurIPS Paper Checklist}

\begin{enumerate}

\item {\bf Claims}
    \item[] Question: Do the main claims made in the abstract and introduction accurately reflect the paper's contributions and scope?
    \item[] Answer: \answerYes{} 
    \item[] Justification: The abstract and introduction describe the paper's scope as cost-aware exploration for LLM agents. We claim that we can induce LLMs to explicitly reason about balancing the cost and uncertainty tradeoffs and act more optimally in the environment by conditioning the LLMs on inferred priors. The claims are supported by the formalization, method, and experimental results in the rest of the paper.

\item {\bf Limitations}
    \item[] Question: Does the paper discuss the limitations of the work performed by the authors?
    \item[] Answer: \answerYes{} 
    \item[] Justification: We discuss the limitations in Appendix \ref{app:limitations}.

\item {\bf Theory assumptions and proofs}
    \item[] Question: For each theoretical result, does the paper provide the full set of assumptions and a complete (and correct) proof?
    \item[] Answer: \answerYes{} 
    \item[] Justification: The paper derives the optimal policy for the discounted Pandora's Box setting and provides the full algorithm and proof in Appendix ~\ref{app:oracle-proof}. The assumptions of the problem are stated in Section \ref{sec:poc}. 

    \item {\bf Experimental result reproducibility}
    \item[] Question: Does the paper fully disclose all the information needed to reproduce the main experimental results of the paper to the extent that it affects the main claims and/or conclusions of the paper (regardless of whether the code and data are provided or not)?
    \item[] Answer: \answerYes{} 
    \item[] Justification: The paper describes the task definitions, datasets, prior estimation procedures, baselines, metrics, and prompt templates (Sec~\ref{sec:prompt-all}) needed to reproduce the main experiments. 

\item {\bf Open access to data and code}
    \item[] Question: Does the paper provide open access to the data and code, with sufficient instructions to faithfully reproduce the main experimental results, as described in supplemental material?
    \item[] Answer: \answerYes{} 
    \item[] Justification: We provide the data used via supplementary materials in OpenReview.

\item {\bf Experimental setting/details}
    \item[] Question: Does the paper specify all the training and test details (e.g., data splits, hyperparameters, how they were chosen, type of optimizer) necessary to understand the results?
    \item[] Answer: \answerYes{} 
\item[] Justification: We specify the details in Sec~\ref{sec:experiment-setup}, Sec~\ref{app:code-data-generation}, Sec~\ref{app:grpo-training-details}.

\item {\bf Experiment statistical significance}
    \item[] Question: Does the paper report error bars suitably and correctly defined or other appropriate information about the statistical significance of the experiments?
    \item[] Answer: \answerYes{} 
    \item[] Justification: We provide significance tests in Sec~\ref{app:sig-testing}.

\item {\bf Experiments compute resources}
    \item[] Question: For each experiment, does the paper provide sufficient information on the computer resources (type of compute workers, memory, time of execution) needed to reproduce the experiments?
    \item[] Answer: \answerYes{} 
    \item[] Justification: We provide details in Sec~\ref{app:grpo-training-details}.
    
\item {\bf Code of ethics}
    \item[] Question: Does the research conducted in the paper conform, in every respect, with the NeurIPS Code of Ethics \url{https://neurips.cc/public/EthicsGuidelines}?
    \item[] Answer: \answerYes{} 
    \item[] Justification: The research uses public datasets, synthetic data, and open model/retriever components for evaluating cost-aware LLM agent behavior. We are not aware of any violations of the NeurIPS Code of Ethics in our work.

\item {\bf Broader impacts}
    \item[] Question: Does the paper discuss both potential positive societal impacts and negative societal impacts of the work performed?
    \item[] Answer: \answerYes{} 
    \item[] Justification: We discuss the impacts in Conclusion section and Sec~\ref{app:impact}.
    
\item {\bf Safeguards}
    \item[] Question: Does the paper describe safeguards that have been put in place for responsible release of data or models that have a high risk for misuse (e.g., pre-trained language models, image generators, or scraped datasets)?
    \item[] Answer: \answerNA{} 
    \item[] Justification: The paper operates on small LLMs and does not directly release a new high-risk model or dataset. We do not perceive the method itself as having a high potential for misuse.

\item {\bf Licenses for existing assets}
    \item[] Question: Are the creators or original owners of assets (e.g., code, data, models), used in the paper, properly credited and are the license and terms of use explicitly mentioned and properly respected?
    \item[] Answer: \answerYes{} 
    \item[] Justification: We have cited relevant papers in the paper.

\item {\bf New assets}
    \item[] Question: Are new assets introduced in the paper well documented and is the documentation provided alongside the assets?
    \item[] Answer: \answerYes{} 
    \item[] Justification: We have documented the construction procedure of the dataset \textsc{FileReading}, design of framework, and eval protocol in the paper. We will also release the code and data.

\item {\bf Crowdsourcing and research with human subjects}
    \item[] Question: For crowdsourcing experiments and research with human subjects, does the paper include the full text of instructions given to participants and screenshots, if applicable, as well as details about compensation (if any)? 
    \item[] Answer: \answerNA{} 
    \item[] Justification: This work does not involve human subjects.

\item {\bf Institutional review board (IRB) approvals or equivalent for research with human subjects}
    \item[] Question: Does the paper describe potential risks incurred by study participants, whether such risks were disclosed to the subjects, and whether Institutional Review Board (IRB) approvals (or an equivalent approval/review based on the requirements of your country or institution) were obtained?
    \item[] Answer: \answerNA{} 
    \item[] Justification: This work does not involve human subjects.

\item {\bf Declaration of LLM usage}
    \item[] Question: Does the paper describe the usage of LLMs if it is an important, original, or non-standard component of the core methods in this research? Note that if the LLM is used only for writing, editing, or formatting purposes and does \emph{not} impact the core methodology, scientific rigor, or originality of the research, declaration is not required.
    \item[] Answer: \answerYes{} 
    \item[] Justification: LLMs are central to the paper's methodology and evaluation. The paper specifies the base model, prompting variants, evaluation settings, and RL-trained variants in main body and the appendix.

\end{enumerate}